%% file: main.tex
\documentclass[10pt,twocolumn,letterpaper]{article}

\usepackage{iccv}
\usepackage{times}
\usepackage{epsfig}
\usepackage{graphicx}
\usepackage{amsmath}
\usepackage{amssymb}
\usepackage{overpic}
\usepackage{wrapfig}
\usepackage{mwe}
\usepackage{color}
\usepackage{adjustbox}
\usepackage{breakcites}

\usepackage[dvipsnames]{xcolor}
\usepackage[colorlinks = true,
            urlcolor  = RubineRed,
          ]{hyperref}

\iccvfinalcopy % *** Uncomment this line for the final submission

\def\iccvPaperID{8223} % *** Enter the ICCV Paper ID here

\usepackage{caption}
\ificcvfinal\fi

\input{symbols}

\input{macros}

\begin{document}

%%%%%%%%% TITLE
\title{ARAPReg: An As-Rigid-As Possible Regularization Loss for Learning Deformable Shape Generators}

\author{Qixing Huang*\\
UT Austin\\
{\tt\small huangqx@cs.utexas.edu}
\and
Xiangru Huang*\\
UT Austin \& MIT\\
{\tt\small xiangruhuang816@gmail.com}
\and
Bo Sun*\\
UT Austin\\
{\tt\small bosun@cs.utexas.edu}
\and
Zaiwei Zhang\\
UT Austin\\
{\tt\small zaiweizhang@utexas.edu}
\and
Junfeng Jiang\\
Hohai University\\
{\tt\small jiangjf.hhu@gmail.com}
\and 
Chandrajit Bajaj \\
UT Austin \\
{\tt\small bajaj@cs.utexas.edu}
}

%\begin{figure*}
%\includegraphics[width=1.0\textwidth]{Figures2/Teaser.pdf}
%\caption{123.}

%\end{figure*}

\twocolumn[{%
\renewcommand\twocolumn[1][]{#1}%
\maketitle
\begin{center}
    \centering
    \vspace{-0.2in}
    \begin{overpic}[width=1.0\textwidth]{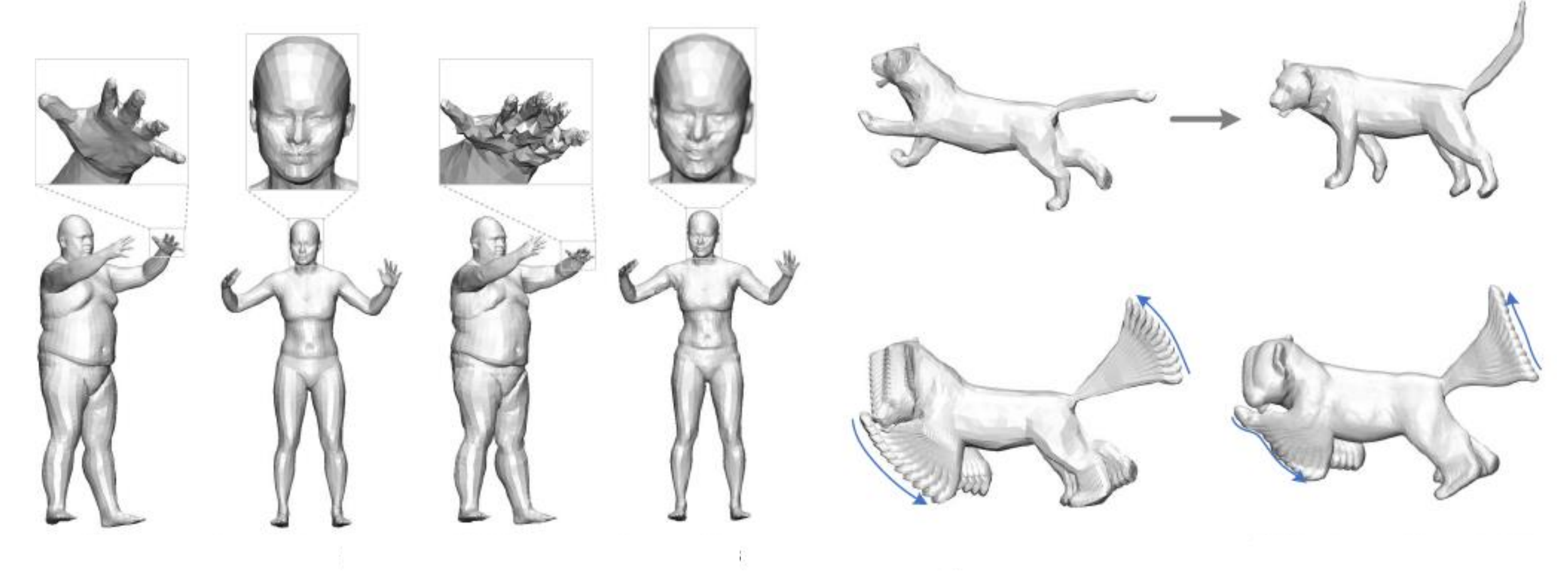}
    \put(8,0){With ARAPReg}
    \put(30,0){Without ARAPReg}
    \put(56,0){With ARAPReg}
    \put(80,0){Without ARAPReg}
    \put(60,20){Source}
    \put(86,20){Target}
    \end{overpic}
\vspace{-0.2in}
    \captionof{figure}{Our approach learns a shape generator from a collection of deformable shapes. The shape generator is trained with a novel as-rigid-as-possible regularization (ARAPReg) loss that promotes the preservation of multi-scale shape features.}
    \label{Figure:Teaser}
\end{center}
%     \begin{center}
%     \centering
%     \vspace{-0.2in}
%     \includegraphics[width=1.0\textwidth]{Figures2/Teaser.pdf}
%     \vspace{-0.2in}
%     \captionof{figure}{We introduce an approach that learns a shape generator from a collection of deformable shapes. The shape generator is trained with regularization terms that promote geometric constraints such as local rigidity and smoothness.}
%     \label{Figure:Teaser}
%     \end{center}%
}]

\ificcvfinal\thispagestyle{empty}\fi

%%%%%%%%% ABSTRACT
\input{00_abstract}

%%%%%%%%% BODY TEXT

\input{01_intro.tex}
\input{02_related.tex}
\input{03_overview.tex}

\input{04_approach.tex}
\input{05_implementation}

\input{06_results.tex}
\input{07_conclusions.tex}

{\small
\bibliographystyle{ieee_fullname}
\bibliography{egbib}
}

\newpage
\clearpage
\appendix
\input{11_supp.tex}
\input{08_app_regu_term.tex}
\input{10_app_gradients}

\end{document}

%% file: symbols.tex
\let \bs = \boldsymbol
\let \set = \mathcal

\newcommand{\R}{\mathcal{R}}

%% file: macros.tex
\let \bs=\mathbf
\let \set=\mathcal

\def \diag {\mathrm{diag}}

\def \saliency {\textup{\saliency}}

\def \path {\mathit{path}}

\newcommand*{\qed}{\hfill\ensuremath{\square}}

\newtheorem{proposition}{\textbf{Proposition}}

\let \set = \mathcal
\let \bs = \boldsymbol

%% file: 00_abstract.tex
\begin{abstract}
This paper introduces an unsupervised loss for training parametric deformation shape generators. The key idea is to enforce the preservation of local rigidity among the generated shapes. Our approach builds on an approximation of the as-rigid-as possible (or ARAP) deformation energy. We show how to develop the unsupervised loss via a spectral decomposition of the Hessian of the ARAP energy. Our loss nicely decouples pose and shape variations through a robust norm. The loss admits simple closed-form expressions. It is easy to train and can be plugged into any standard generation models, e.g., variational auto-encoder (VAE) and auto-decoder (AD). Experimental results show that our approach outperforms existing shape generation approaches considerably on public benchmark datasets of various shape categories such as human, animal and bone. Our code and data are available at  \href{https://github.com/GitBoSun/ARAPReg}{https://github.com/GitBoSun/ARAPReg}.
%datasets such as DFAUST, SMAL, and Bone.
\end{abstract}

%% file: 01_intro.tex
\section{Introduction}
\label{Section:Introduction}

This paper considers learning a parametric mesh generator from a deformable shape collection with shapes that exhibit the same topology but undergo large geometric variations (see examples below of a deforming human, animal, and bone). This problem arises in numerous visual computing and relevant fields such as recovery of neural morphogenesis, data-driven shape reconstruction, and image-based reconstruction, to name just a few (c.f.~\cite{Xu:2016:Survey}). 

Deformable shapes differ from many other visual objects (e.g., images and videos) because there are natural constraints  underlying the shape space. One such example is the local rigidity constraint; namely, corresponding surface patches among neighboring shapes in the shape space undergo approximately rigid transformations. This constraint manifests the preservation of geometric features (e.g., facial features of humans and toes of animals) among local neighborhoods of the underlying shape space. An interesting problem thus is the use of this constraint to train shape generators from a collection of training shapes, where the local rigidity constraint accurately and efficiently propagates features of the training shapes to new synthetic shapes produced by the generator.  

%When applying to train shape generators, it  Despite the importance of this constraint, its application to training shape generators is significantly under-explored. 

% \begin{figure}[t]
% \centering
% \begin{overpic}[width=1.0\columnwidth]{Figures2/Teaser_half.pdf}
% \end{overpic}
% \vspace{0.05in}
% \caption{Shape generator trained with as-rigid-as-possible (ARAP) regularization terms exhibits better local rigidity and smoothness.}
% \label{Figure:Teaser}
% \end{figure}

In this paper, we study how to model the local rigidity constraint as an unsupervised loss functional for generative modeling. The proposed loss can be combined with standard mesh generators such as variational auto-encoders (VAEs)~\cite{Tan0LX18,LitanyBBM18,ranjan2018generating,bouritsas2019neural} and auto-decoders (ADs)~\cite{YangLH18,zadeh2019variational}. A key property of our loss functional is that it is consistent with other training losses. This property offers multiple advantages. For example, the learned generator is insensitive to the tradeoff parameters among the loss terms. As another example, the training procedure converges faster than the setting where loss terms may compete against each other.

Our approach, called ARAPReg, builds on the established as-rigid-as-possible (or ARAP) deformation model~\cite{Sorkine:2007:ARAP,WandJHBGS07,Xu:2007:GDE} that measures the non-rigid deformation between two shapes. Its key ingredients include use oof the Hessian of the ARAP deformation model to derive an explicit regularizer for the Jacobian of the shape  generator and a robust norm on the Hessian to model pose and shape variations of deformable shapes. The outcome is a simple closed-form formulation for training mesh generators. ARAPReg differs from prior works that enforce ARAP losses between synthetic shapes and a base shape~\cite{DBLP:conf/cvpr/HabermannXZPT20,DBLP:conf/nips/LiLMKW0K20,ZhouBP20}, that may introduce competing losses when the underlying shape space has large deformations. 

We have evaluated ARAPReg across a variety of public benchmark datasets such as DFAUST~\cite{dfaust:CVPR:2017}, SMAL~\cite{zuffi20173d}, and an in-house benchmark dataset of Bone. The evaluations include both generator settings of VAE and AD. Experimental results show that ARAPReg leads to considerable performance gains across state-of-the-art deformable shape generators both qualitatively and quantitatively. As shown in Figure~\ref{Figure:Teaser} for example, the interpolated shapes using ARAPReg greatly preserve the local geometric details of the generated shapes and avoids unrealistic shape poses.

%% file: 02_related.tex
\section{Related Works}
\label{Section:Related:Works}

This section organizes the relevant works into three groups, namely, 3D generative models, regularization for generative modeling, and shape space modeling.
%and distance metrics for non-rigid shape matching. 

%\subsection{Parametric Generative Models}
%\label{Subsec:Parametric:Models}

\noindent\textbf{3D generative models.} 
%Methods for learning parametric models differ from each other in representing the visual data and the mapping function. Early works on learning parametric models focus on faces and human bodies~\cite{Blanz:1999:MMS,Allen:2003:SHB,Anguelov:2005:SSC}. The parametric models are given by linearly blending exemplar models in a model collection. Such a method is only applicable to object classes with small geometric variability. However, they do not generalize well to shape collections that exhibit significant variations. ARAPReg addresses this issue by learning a non-linear generator that explicitly enforces geometric constraints to enhance the plausibility of the generated models.
%Stimulated from recent advances in deep models for generating visual data such as GAN~\cite{GoodfellowPMXWOCB14}, VAE~\cite{Kingma:2014:VAE}, and their variants~\cite{Radford:ICLR:2016,IsolaZZE17,ZhuPIE17,GulrajaniAADC17,ArjovskyCB17}, there have been extensive works on extending these machine learning techniques to 3D data. 
Learning 3D generative models relies on developing suitable 3D representations to encode 3D models into vectorized forms. Examples include volumetric grid~\cite{NIPS2016_6096,TatarchenkoDB17,DaiQN17,ParkYYCB17,MeschederONNG19,HeGDG20}, implicit surfaces~\cite{ParkFSNL19,DBLP:conf/cvpr/ChenZ19}, point clouds~\cite{AchlioptasDMG18,YangFST18,YangHH0BH19,LiZZPS19}, meshes~\cite{KatoUH18,LitanyBBM18,10.1145/3355089.3356488}, parametric surfaces~\cite{GroueixFKRA18,Maron:2017:STC}, spherical representations~\cite{DBLP:journals/corr/abs-1801-10130,DBLP:journals/corr/abs-1711-06721,DBLP:journals/corr/abs-1712-04426}, geometric arrangements~\cite{Tulsiani2017Abstractions,Zhang:2018:SG}, and multi-views~\cite{DBLP:conf/3dim/LunGKMW17}. 

This paper is mostly relevant to generative models under the mesh representation, which falls into four categories. The first category of approaches~\cite{Tan0LX18,DBLP:conf/cvpr/VermaBV18,LitanyBBM18,TretschkTZGT20,RakotosaonaO20} is based on defining variational auto-encoders on meshes. A typical strategy is to treat triangular meshes as graphs and define convolution and deconvolution operations to synthesize triangular meshes (c.f.~\cite{DBLP:conf/cvpr/VermaBV18,LitanyBBM18,TretschkTZGT20}). \cite{Tan0LX18} introduced a geometric encoding scheme that operates in the gradient domain. The second category of approaches builds upon recurrent procedures for geometric synthesis. This methodology has been extensively applied for primitive-based assembly~\cite{Li:2017:GGR,Ritchie:2016:NPM,DBLP:journals/corr/abs-1712-08290,DBLP:conf/iccv/ZouYYCH17}. \cite{Hanocka:2019:MCNN} extended this approach to meshes, in which edge contraction operations are applied recursively. The third category of approaches~\cite{YumerM16,DBLP:conf/cvpr/WangAKCS20} deforms a base mesh to generate new meshes, where the deformation is learned from data. The last category utilizes surface parameterization~\cite{SinhaUHR17,Maron:2017:STC,GroueixFKRA18,Hamu:2018:Multichart}. 

While these approaches focused on adopting generative modeling methodologies under the mesh setting, ARAPReg studies the novel problem of explicitly enforcing an ARAP loss among synthetic shapes with similar latent codes. 

%\subsection{3D Representations for Mesh Generation}
%\label{Subsec:3D:Representations}

\noindent\textbf{Regularization for generative modeling.} Regularization losses have been explored in prior works for 3D generative modeling. In~\cite{DBLP:conf/eccv/PeeblesPZE020}, Peebles et al. studied a Hessian regularization term for learning generative image models. A spectral regularization loss is introduced in~\cite{DBLP:conf/iccv/Aumentado-Armstrong19} for 3D generative modeling. Several works~\cite{DBLP:journals/ijcv/WuXLTTTF18,DBLP:conf/eccv/WangZLFLJ18,DBLP:conf/cvpr/KanazawaBJM18,DBLP:conf/3dim/BalashovaSWTCF18} studied geometric regularizations for image-based reconstruction. In contrast, ARAPReg focuses on regularization terms that are consistent with other terms. Several other works~\cite{DBLP:conf/cvpr/HabermannXZPT20,DBLP:conf/nips/LiLMKW0K20,ZhouBP20} employed ARAP losses between any synthetic shapes with a base shape. The novelty of ARAPReg is that it is consistent with other loss terms even when the underlying shape space presents large deformations. The reason is that the local rigidity constraint is only enforced among neighboring shapes in the underlying shape space. Our initial experiments show that enforcing ARAP losses between synthetic shapes and a base shape leads to worse results than dropping the ARAP losses.

\noindent\textbf{Shape space modeling.} Finally, ARAPReg is relevant to early works on modeling tangent spaces of shape manifolds~\cite{Kilian:2007:GMS,HuangWAG09,Yang:2011:SSE}. However, unlike the applications in shape interpolation~\cite{Kilian:2007:GMS}, shape segmentation~\cite{HuangWAG09}, and mesh-based geometric design~\cite{10.1145/1778765.1778780,Yang:2011:SSE}, ARAPReg focuses on devising an unsupervised loss for network training. 

%% file: 03_overview.tex
\section{Overview}
\label{Section:Overview}

\begin{figure}[h]
\begin{overpic}[width=1.0\columnwidth]{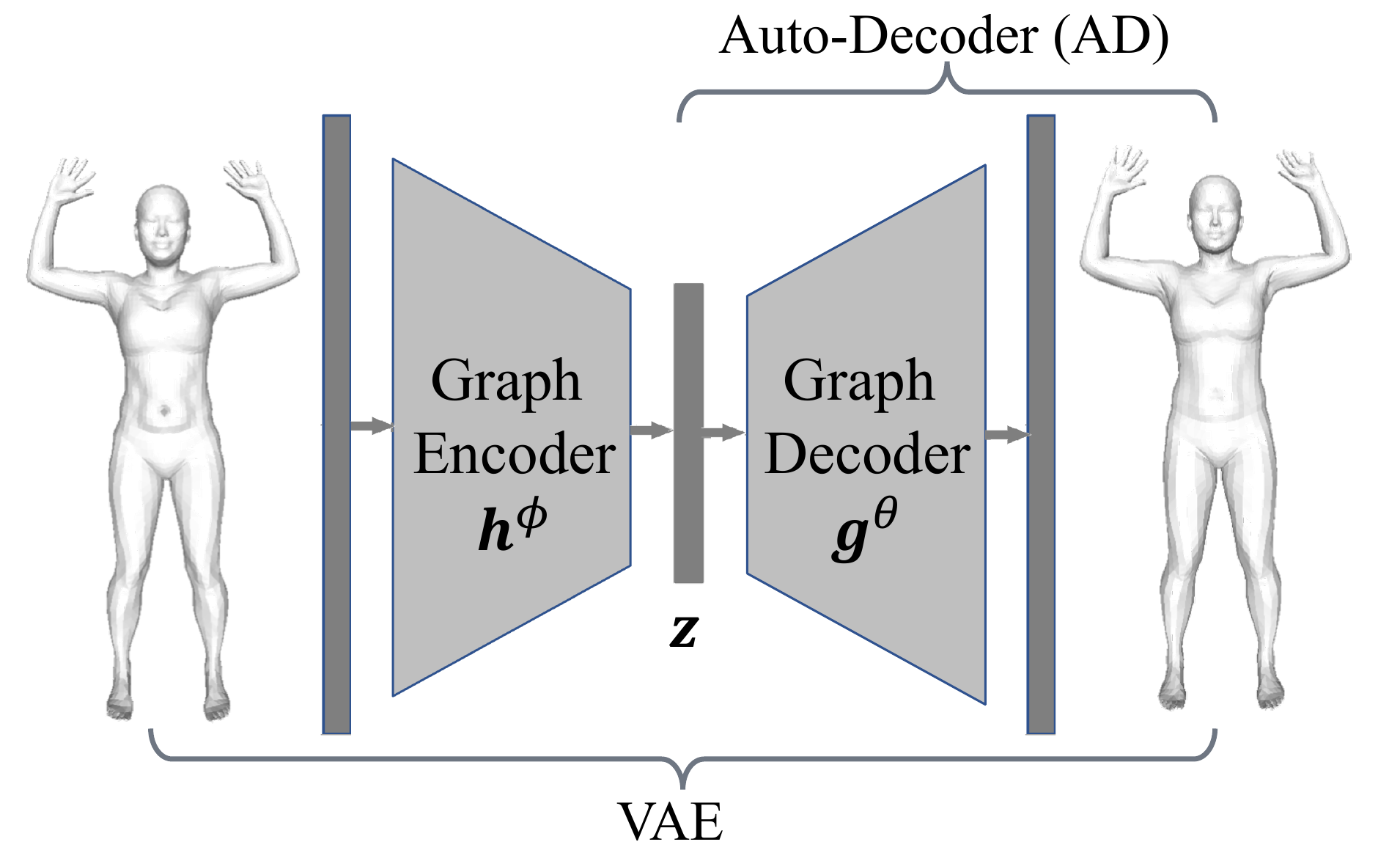}
% \put(0,6){$\bs{z}$}
% \put(20,-4.5){$d\bs{p}^{\theta}(\bs{z})$}
% \put(32,16.7) {$\bigoplus$}
% \put(43,-4.5) {Base mesh $\bs{p}^{\textup{base}}$}
% \put(80,-4.5) {Output $\bs{g}^{\theta}(\bs{z})$}
% \put(8.5, 21) {\small{Graph}}
% \put(7.5, 16.5) {\small{Deconv.}}
% \put(12, 11) {$\theta$}
\end{overpic}
\vspace{-0.1in}
\caption{We consider multiple standard shape generators, including Variational Auto-Encoder (VAE) and Auto-Decoder (AD). The graph encoder $\bs{h}^{\phi}$ maps the input mesh $\bs{g}$ to a latent parameter $\bs{h}^{\phi}(\bs{z})$. The graph decoder maps a latent parameter $\bs{z}$ to the out mesh $\bs{g}^{\theta}(\bs{z})$. }
\vspace{-0.2in}
\label{Figure:Base:Mesh:Deformation}
\end{figure}

Following~\cite{ranjan2018generating,bouritsas2019neural,zhou2020fully}, we are interested in learning a mesh generator that takes a latent code as input and outputs the vertex positions of a triangular mesh with given mesh connectivity (See Figure~\ref{Figure:Base:Mesh:Deformation}). Formally speaking, we denote this mesh generator as 
$$
\bs{g}^{\theta}: \set{Z}:=\R^k \rightarrow \R^{3n}.
$$
Here $\set{Z}$ represents the latent space, and $\R^{3n}$ encodes the vector that concatenates the vertex positions, i.e., $n$ is the number of vertices. We organize the remainder of this paper as follows.

In Section~\ref{Section:Approach}, we introduce the key contribution of this paper, ARAPReg, an unsupervised loss for training $\bs{g}^{\theta}$. The loss only requires a prior distribution of the latent space $\set{Z}$. In this paper, we assume the prior distribution is the Normal distribution $\set{N}_k$ of dimension $k$. The key idea of ARAPReg is to ensure that the local rigidity constraint is preserved among neighboring generated shapes in the underlying shape space. As illustrated in the left part of Figure~\ref{Figure:Teaser}, the goal of this loss is to significantly improve the generalization behavior of the mesh generator, e.g., preserving multi-scale geometric details. 

%Moreover, ARAPReg nicely decouples shape and pose variations among the generated shapes through a simple formulation. 

In Section~\ref{Section:Implementation:Details}, we discuss how to plug this unsupervised loss into standard shape generation models based on VAE and AD. 

%We also discuss the implementation details. 

%% file: 04_approach.tex
\section{Formulation of the ARAPReg Loss}
\label{Section:Approach}

Formulating the preservation of local rigidity is quite challenging because the resulting loss term has to be simple enough to facilitate network training.  One straightforward approach is to enforce the local rigidity constraint between a generated shape $\bs{g}^{\theta}(\bs{z})$ and its perturbation $\bs{g}^{\theta}(\bs{z}+d\bs{z})$. Here $d\bs{z}$ characterizes an infinitesimal displacement in the parameter space. However, this approach requires sampling a lot of shape pairs. Besides, typical formulations of shape deformations between $\bs{g}^{\theta}(\bs{z})$ and $\bs{g}^{\theta}(\bs{z}+d\bs{z})$ require solving optimization problems that are computationally expensive (c.f.~\cite{DeformationSurvey:2008}).

ARAPReg stitches several novel ideas to derive a simple unsupervised loss that does not adversely compete with typical losses used in generative modeling (See Section~\ref{Subsec:Arch}). 

\subsection{Step I: Decoupling Smoothness and Jacobian regularization} 

First, ARAPReg decouples the enforcement of local rigidity into two terms. The first term enforces the smoothness of the generator. This smoothness penalty enables the second term, which formulates the preservation of local rigidity as potentials on the Jacobian of the generator, i.e.,  
$$
\frac{\partial \bs{g}^{\theta}}{\partial \bs{z}}(\bs{z}) \in \R^{(3n)\times k}.
$$

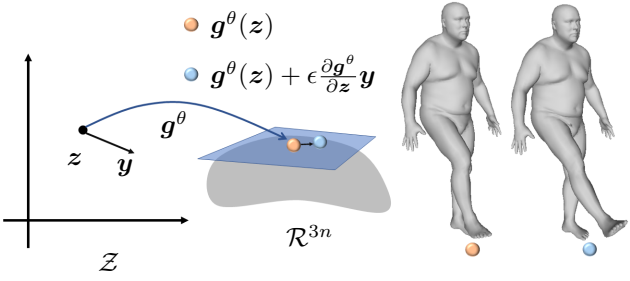
\begin{figure}
\centering
\begin{overpic}[width=1.0\columnwidth]{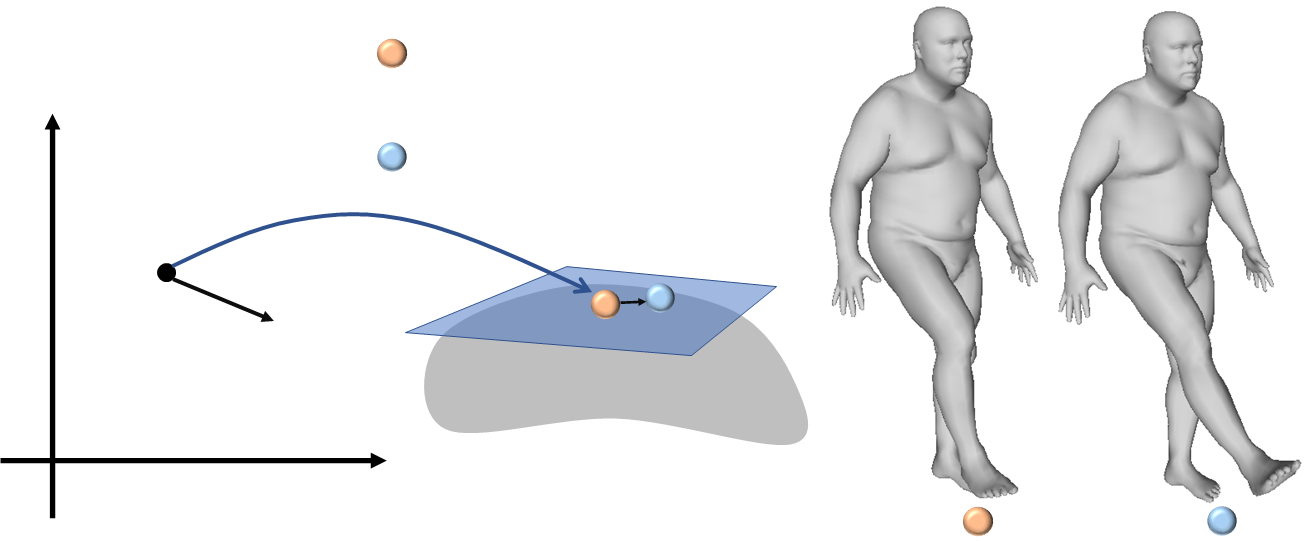}
\put(15,-2){$\set{Z}$}
\put(18,14){$\bs{y}$}
\put(25,20){$\bs{g}^{\theta}$}
\put(10,15){$\bs{z}$}
\put(45,2){$\R^{3n}$}
\put(33,36){$\bs{g}^{\theta}(\bs{z})$}
\put(33,28){$\bs{g}^{\theta}(\bs{z})+\epsilon \frac{\partial \bs{g}^{\theta}}{\partial \bs{z}} \bs{y}$}
\vspace{0.4in}
\end{overpic}
\caption{Illustration of configuration for Jacobian regularization. We study infinitesimal deformations incurred by the tangent space at each generated shape $\bs{g}^{\theta}(\bs{z})$.}
\label{Figure:Tangent:Space:Illus}
\vspace{-0.2in}
\end{figure}

Specifically, we define the unsupervised loss as
\begin{align}
& \set{L}_{reg}(\theta) := \underset{\bs{z}\sim \set{N}_k}{E}\Big(\underset{\delta\bs{z}\sim s\set{N}_k}{E}\|\bs{g}^{\theta}( \bs{z}+\delta\bs{z})-2\bs{g}^{\theta}(\bs{z}) + \nonumber \\
&\qquad\quad  \bs{g}^{\theta}( \bs{z}-\delta\bs{z})\|^2 + \lambda_{R}\cdot r_{R}(\bs{g}^{\theta}(\bs{z}),\frac{\partial \bs{g}^{\theta}}{\partial \bs{z}}(\bs{z}))\Big), 
\label{Eq:Regu:Total:Regu:Loss}
\end{align}
where the first term promotes the smoothness of the generator $\bs{g}^{\theta}$; $s$ is a hyper-parameter of ARAPReg. Note that unlike enforcing 
\begin{equation}
\bs{g}^{\theta}( \bs{z}+\delta\bs{z})\approx \bs{g}^{\theta}( \bs{z}) + \frac{\partial \bs{g}^{\theta}}{\partial \bs{z}}(\bs{z})\cdot \delta \bs{z},
\label{Figure:Smoothness:Cons}
\end{equation}
the formulation in (\ref{Eq:Regu:Total:Regu:Loss}) does not involve the first-order derivatives of $\bs{g}$. It follows that network training is more efficient as it only requires computing the first-order derivatives of $\bs{g}$. On the other hand, it penalizes the second-order derivatives of $\bs{g}^{\theta}$. It therefore implicitly enforces (\ref{Figure:Smoothness:Cons}). The second term $r_{R}(\bs{g}^{\theta}(\bs{z}),\frac{\partial \bs{g}^{\theta}}{\partial \bs{z}}(\bs{z}))$ in (\ref{Eq:Regu:Total:Regu:Loss}), which will be defined shortly,  formulates the regularization loss concerning the generated mesh $\bs{g}^{\theta}(\bs{z})$ and infinitesimal perturbations specified by the Jacobian $\frac{\partial \bs{g}^{\theta}}{\partial \bs{z}}(\bs{z})$ (See Figure~\ref{Figure:Tangent:Space:Illus}).  $\lambda_{R}$ is another hyper-parameter of ARAPReg. 

In other words, instead of enforcing the local rigidity between shape pairs, ARAPReg enforces the preservation of the local rigidity in the tangent space specified by the Jacobian. The tangent space is a first-order approximation of the shape space. The smoothness potential ensures that this first-order approximation is accurate, i.e., the rigidity constraint propagates to the shape space's local neighborhood. As we will discuss later, another appealing property of this formulation is that the Jacobian enables us to easily model pose and shape variations (where pose variations are more rigid than shape variations). This goal is hard to achieve using generic pairwise regularizations. 

Although the smoothness constraint involves shape pairs, our experiments suggest that there is no need to sample a large number of shape pairs. One interpretation is that deep neural network training has implicit regularizations (c.f.~\cite{NeyshaburTSS17}), which promotes smoothness. 

\subsection{Step II: Jacobian Regularization} 

We proceed to introduce the local rigidity term $r_{R}$ that regularizes the Jacobian of the generator. To make the notations uncluttered, we focus on formulating $r_{R}(\bs{g},J)$. Here $\bs{g}\in \R^{3n}$ denotes a vertex position vector, and $J\in \R^{3n\times k}$ is a Jacobian matrix that specifies infinitesimal perturbations to $\bs{g}$. 

Our formulation is inspired by the as-rigid-as possible (or ARAP) potential function~\cite{Sorkine:2007:ARAP,WandJHBGS07,Xu:2007:GDE}. This standard model measures the deformation between a pair of shapes. Consider a mesh with vertex position $\bs{g} \in \R^{3n}$ and the same mesh with perturbed vertex position $\bs{g}+\bs{x}\in \R^{3n}$. Denote $O_i\in SO(3)$ as the latent rotation associated with the $i$-th vertex. The ARAP deformation between them is 
\begin{align}
& f_{R}(\bs{g},\bs{x}) :=  \min\limits_{O_i\in SO(3)}\sum\limits_{(i,j)\in \set{E}}\|\bs{r}_{ij}(O_i,\bs{g},\bs{x})\|^2 \label{Eq:ARAP:Def}    \\
& \bs{r}_{ij}(O_i,\bs{g},\bs{x}):=(O_i-I_3)(\bs{g}_i-\bs{g}_j)-\big(\bs{x}_i-\bs{x}_j\big)
\nonumber  
\end{align} 
where $\set{E}$ denotes the edge set of the mesh generator $\bs{g}^{\theta}$. Note that we assume $(i,j)\in \set{E}$ if and only if $(j,i)\in \set{E}$. 

To introduce a formulation that only depends on the Jacobian of the generator, we consider the Taylor expansion of the as-rigid-as possible potential energy.

\begin{proposition} (\cite{HuangWAG09})
The zero and first-order derivatives of $f_{R}$ satisfy
\begin{align*}
f_{R}(\bs{g},\bs{0}) = 0,\qquad 
\frac{\partial f_{R}}{\partial \bs{x}}(\bs{g},\bs{0}) =\bs{0}. 
\end{align*}
Moreover, the Hessian matrix is given by 
\begin{align}
\frac{\partial^2 f_{R}}{\partial^2 \bs{x}}(\bs{g},\bs{0}) & = H_{R}(\bs{g}), \nonumber \\
H_{R}(\bs{g}) & = L\otimes I_3 - A(\bs{g})^T D(\bs{g})^{-1} A(\bs{g}),
\label{Eq:ARAP:2:Order}
\end{align}
where $L\in \R^{n\times n}$ is the graph Laplacian associated to $\set{E}$; $A(\bs{g})$ is a sparse $n\times n$ block matrix; $D(\bs{g})$ is a diagonal block matrix. The blocks of $A(\bs{g})$ and $D(\bs{g})$ are given by
\begin{align*}
A_{ij}(\bs{g}) & = \left\{
\begin{array}{cc}
\sum\limits_{k\in \set{N}(i)}(\bs{v}_{ik}\times) & \qquad\qquad\quad \ \ i=j \\
-\bs{v}_{ij}\times & \ \ \qquad\qquad\quad (i,j)\in \set{E} \\
0 &\qquad\qquad\quad \ \  \textup{otherwise}
\end{array}
\right.\ \\
D_{ij}(\bs{g}) & = \left\{
\begin{array}{cc}
\sum\limits_{k\in \set{N}(i)}(\|\bs{v}_{ik}\|^2 I_3 -\bs{v}_{ik}\bs{v}_{ik}^T) & i = j \\
0 & \textup{otherwise}
\end{array}
\right.\
\end{align*}
where $\bs{v}_{ij} = \bs{g}_i-\bs{g}_j$, and $\set{N}(i)$ collects indices of adjacent vertices of $i$. 
Note that $H_R$ is a highly sparse matrix. 
\label{Prop:Rigidity}
\end{proposition}

%\noindent\textsl{Proof:} Please refer to the supp. material. \qed

Proposition~\ref{Prop:Rigidity} indicates that for each vector $\bs{y}\in \R^d$ in the parameter space, the ARAP potential between the mesh defined by $\bs{g}$ and its infinitesimal displacement encoded by $\epsilon J\bs{y}$ (for a small $\epsilon$) can be approximated as
\begin{equation}
f_{R}(\bs{g}, \epsilon J \bs{y})\approx \frac{1}{2}\epsilon^2 \bs{y}^T\overline{H}_{R}(\bs{g},J)\bs{y},
\label{Eq:F:R:Approx}
\end{equation}
where $\overline{H}_{R}(\bs{g},J):=J^TH_{R}(\bs{g})J$.

\begin{figure*}
\centering
\includegraphics[width=0.9\textwidth]{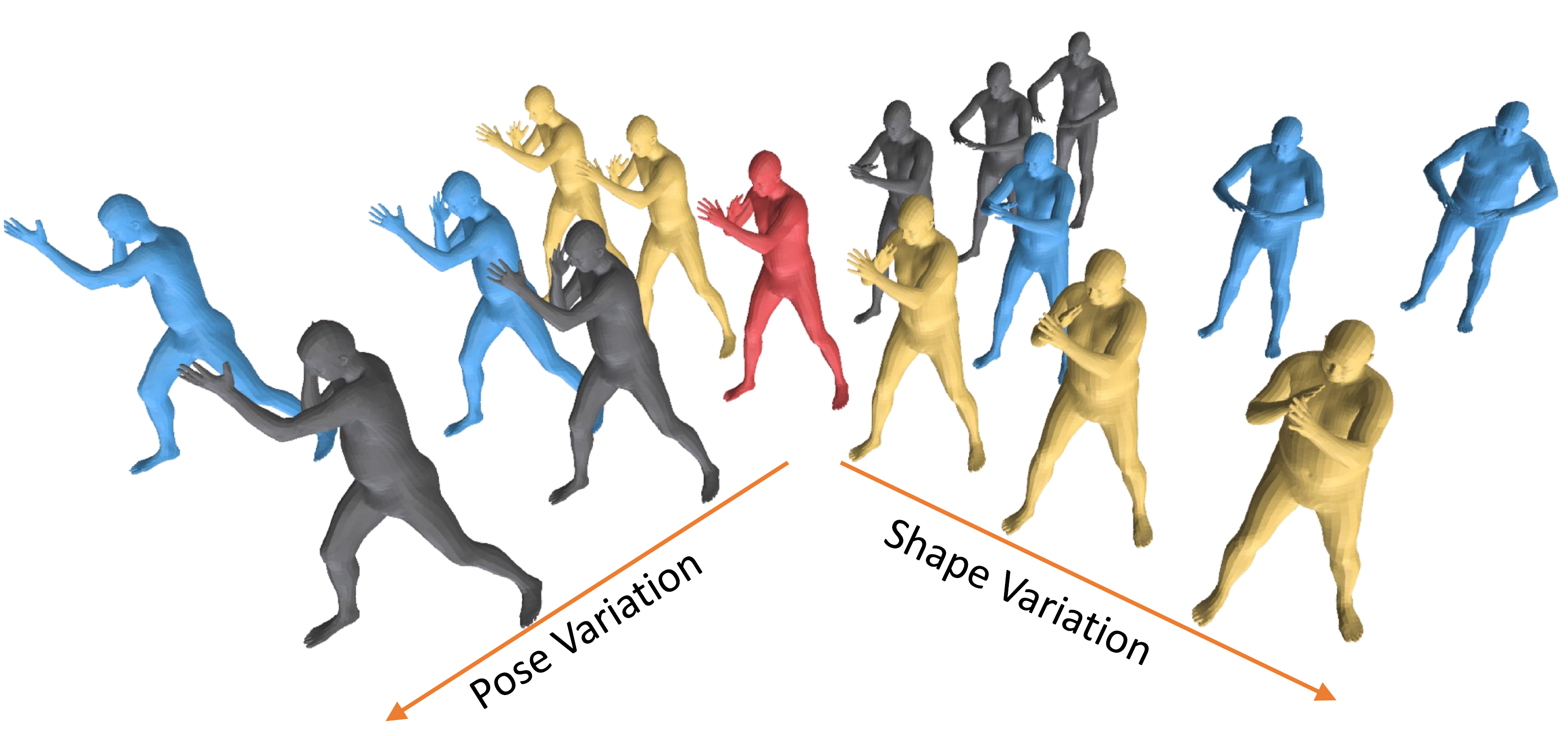}
\vspace{-0.1in}
\caption{This figure illustrates the local shape space spanned by the  eigenvectors of the Hessian of $\overline{H}_{R}(\bs{g},J)$. The red shape in the center is the reference shape. When moving the latent parameter along the first eigenvector, the shape deformation is locally rigid, exhibiting pose variations (see grey shapes). When moving along the largest eigenvector, the shape deformation possesses local stretching, corresponding to shape variations (see yellow shapes). Finally, when moving along a linear combination of both eigenvectors, the shape exhibits both pose and shape variations (see blue shapes).}
\vspace{-0.1in}
\label{Figure:Tangent:Space:Illustration}
\end{figure*}

(\ref{Eq:F:R:Approx}) provides the rigidity potential along a direction $\bs{y}$ in the latent space. Our formulation of $r_R$ seeks to integrate (\ref{Eq:F:R:Approx}) over all possible directions $\bs{y}$. To motivate the final formulation of ARAPReg, let us first define an initial potential energy by integrating $\bs{y}^T\overline{H}_{R}(\bs{g},J)\bs{y}$ over the unit-sphere $\set{S}^k$ in $\R^k$ that specifies all possible $\bs{y}$:
\begin{align}
r_{R}^{L^2}(\bs{g}, J) & := \frac{k}{\textup{Vol}(\set{S}^k)}\int_{\bs{y}\in \set{S}^k} \bs{y}^T\overline{H}_{R}(\bs{g},J)\bs{y} d\bs{y}. 
\label{Eq:ARAP:Regu:Def}
\end{align}

\begin{proposition}
\begin{equation}
r_{R}^{L^2}(\bs{g}, J) = \textup{Tr}(\overline{H}_{R}(\bs{g},J)) = \sum\limits_{i=1}^{k}\lambda_i(\overline{H}_{R}(\bs{g},J))
\label{Eq:ARAP:Regu2}
\end{equation}
where $\lambda_i(\overline{H}_{R}(\bs{g},J))$ is the $i$-th eigenvalue of $\overline{H}_{R}(\bs{g},J)$.
\label{Prop:Local:Rigidity:Expression}
\end{proposition}
%\vspace{-0.1in}
%\noindent\textsl{Proof:} Please refer to the supp. material. \qed

\subsection{Step III: Pose and Shape Variation Modeling}

We present a simple formulation that decouples enforcing pose and shape variations. Specifically, the eigenvalues $\lambda_i(\overline{H}_{R}(\bs{g},J)) = \bs{u}_i^T\overline{H}_{R}(\bs{g},J)\bs{u}_i$, where $\bs{u}_i$ is the corresponding eigenvector of $\lambda_i(\overline{H}_{R}(\bs{g},J))$, reveal the deformations in different directions of the tangent space. From the definition of the as-rigid-as possible deformation energy, each vertex's one-ring neighborhood is mostly rigid under pose variations. In contrast, the one-ring neighborhoods may change drastically under shape variations. This means eigenvectors with small eigenvalues correspond to pose variations, while eigenvectors with large eigenvalues correspond to shape variations (See Figure~\ref{Figure:Tangent:Space:Illustration}).

The limitation of the L2 formulation described in (\ref{Prop:Local:Rigidity:Expression}) is that all directions are penalized equally. ARAPReg employs a robust norm to model the local rigidity loss to address this issue
\begin{equation}
r_R(\bs{g}, J) = \sum\limits_{i=1}^{k}\lambda_i^{\alpha}(\overline{H}_{R}(\bs{g},J)),
\label{Eq:ARAP:Regu}     
\end{equation}
where we set $\alpha = \frac{1}{2}$ in this paper. Similar to the effects of using robust norms for outlier removal, (\ref{Eq:ARAP:Regu}) imposes small weights on the subspace spanned by eigenvectors of large eigenvalues, which correspond to shape variations. In other words, minimizing (\ref{Eq:ARAP:Regu}) minimizes the small eigenvalues of $\overline{H}_{R}(\bs{g},J)$ automatically, which correspond to pose variations. Note that several prior works~\cite{ZhouBP20,DBLP:conf/eccv/CosmoNHKR20,DBLP:conf/iccv/Aumentado-Armstrong19} aimed to decouple pose and shape in the latent space. In contrast, our goal is to model the regularization term by taking pose and shape variations into account. 

%In the meantime,  it will leave a few large eigenvalues of $\overline{H}_{R}(\bs{g},J)$, which correspond to shape variations.

\subsection{Step IV: Final Loss Term} 

Substituting (\ref{Eq:ARAP:Regu:Def}) into (\ref{Eq:Regu:Total:Regu:Loss}), we have
\begin{align}
& \set{L}_{reg}(\theta) :=  \ \underset{\bs{z}\sim \set{N}_k}{E}\Big(\underset{\delta\bs{z}\sim s\set{N}_k}{E}\|\bs{g}^{\theta}( \bs{z}+\delta\bs{z})-2\bs{g}^{\theta}(\bs{z})\nonumber  \\
&\ + \bs{g}^{\theta}( \bs{z}-\delta\bs{z})\|^2 + \lambda_{R}\sum\limits_{i=1}^{k}\lambda_i^{\alpha}\big(\overline{H}_R\big(\bs{g}^{\theta}(\bs{z}),\frac{\partial \bs{g}^{\theta}}{\partial \bs{z}}(\bs{z})\big)\big)\Big) 
\label{Eq:Regu:Total:Regu:Loss2}
\end{align}
In this paper, we set $s = 0.05$ and $\lambda_R = 1$ for all of our experiments. 

The major challenge of using (\ref{Eq:Regu:Total:Regu:Loss2}) for training is to compute the gradient of the Jacobian regularization term. Similar to the formulation of generator smoothness, we introduce a gradient computation approach that only requires computing the derivatives of $\bs{g}^{\theta}$. Please refer to the supp. material for details.

%% file: 05_implementation.tex
\section{Application in Learning Mesh Generators}
\label{Section:Implementation:Details}

This section introduces the applications of the unsupervised loss for learning mesh generators. We first introduce the network architecture used in this paper for experimental evaluation. We then introduce how to insert the unsupervised loss $\set{L}_{reg}$ described above into two formulations of training mesh generators, i.e., variational auto-encoders~\cite{LitanyBBM18,ranjan2018generating,bouritsas2019neural} and auto-decoders~\cite{YangLH18,zadeh2019variational}.

\subsection{Network Architecture}
\label{Subsec:Arch}

%\begin{figure}
%\centering
%\vspace{1in}
%\caption{This figure illustrates the network used for experimental evaluation. We only show the decoder. The encoder is the mirror of the decoder.}
%\label{Figure:Arch}
%\end{figure}

We focus on describing the decoder network $\bs{g}^{\theta}$. When training variational auto-encoders, we utilize another encoder network $\bs{h}^{\phi}:\R^{3n}\rightarrow \set{Z}$, the mirror of $\bs{g}^{\theta}$ but has different network weights. In other words, $\bs{h}^{\phi}$ has the identical network layers as $\bs{g}^{\theta}$, but the connections are reversed.

In this paper, we model $\bs{g}^{\theta}$ using six layers. The second to the sixth layers are the same as the network architecture of~\cite{LitanyBBM18}. Motivated from~\cite{zhou2020fully}, we let the first layer concatenate the latent features associated with each vertex of the coarse mesh as input. Between the input and the first activation is a fully connected layer. Please refer to the supp. material for details.

\subsection{Variational Auto-Encoder}
\label{Subsec:VAE}

Given a collection of training meshes $\set{T} = \{\bs{g}_i|1\leq i \leq N\}$, we solve the following optimization problem to train the auto-encoder that combines $\bs{g}^{\theta}$ and $\bs{h}^{\phi}$:
\begin{align}
\min\limits_{\theta, \phi} & \frac{1}{N}\sum\limits_{i=1}^{N} \|g^{\theta}(h^{\phi}(\bs{g}_i)) - \bs{g}_i\| + \lambda_{KL} KL(\{h^{\phi}(\bs{g}_i)\}|\set{N}_k) \nonumber \\
& + \lambda_{reg}\set{L}_{reg}(\theta)
\label{Eq:VAE:Obj}
\end{align}
where the first two terms of (\ref{Eq:VAE:Obj}) form the standard VAE loss. In this paper, we set $\lambda_{KL} = 1$ and $\lambda_{reg} = 10$. For network training, we employ ADAM~\cite{KingmaB14}.

\subsection{Auto-Decoder}
\label{Subsec:GAN}

The auto-decoder formulation~\cite{YangLH18,zadeh2019variational} replaces the encoder with latent variables $\bs{z}_i$ associated with the training meshes:
\begin{align}
\min\limits_{\theta, \{\bs{z}_i\}} & \frac{1}{N}\sum\limits_{i=1}^{N} \|\bs{g}^{\theta}(\bs{z}_i) - \bs{g}_i\| + \lambda_{KL} KL(\{\bs{z}_i\}|\set{N}_k) \nonumber \\
& + \lambda_{reg}\set{L}_{reg}(\theta)
\label{Eq:Theta:Z}
\end{align}
where we use the same hyper-parameters as Section~\ref{Subsec:VAE}. 

We apply alternating minimization to solve (\ref{Eq:Theta:Z}). The latent parameters $\bs{z}_i$ are initialized as an empirical distribution of $\set{N}_k$. When $\bs{z}_i$ are fixed, (\ref{Eq:Theta:Z}) reduces to
\begin{equation}
\min\limits_{\theta} \frac{1}{N}\sum\limits_{i=1}^{N} \|\bs{g}^{\theta}(\bs{z}_i) - \bs{g}_i\| + \lambda_{reg}\set{L}_{reg}(\theta)
\label{Eq:Theta:Opt}
\end{equation}
We again employ ADAM~\cite{KingmaB14} to solve~(\ref{Eq:Theta:Opt}). Our implementation applies one epoch of optimizing $\theta$ for each alternating optimization iteration. 

When the network parameters $\theta$ are fixed, (\ref{Eq:Theta:Z}) reduces to
\begin{equation}
\min\limits_{\{\bs{z}_i\}} \frac{1}{N}\sum\limits_{i=1}^{N} \|\bs{g}^{\theta}(\bs{z}_i) - \bs{g}_i\| + \lambda_{KL} KL(\{\bs{z}_i\}|\set{N}_k)
\label{Eq:Z:Opt}
\end{equation}
We again employ ADAM~\cite{KingmaB14} to optimize $\bs{z}_i$. Similarly, our implementation applies one epoch of optimizing $\bs{z}_i$ for each alternating iteration. The total number of alternating iterations is set as $30$ in this paper.

%% file: 06_results.tex
\section{Experimental Evaluation}
\label{Section:Results}
\begin{figure*}
\centering
\begin{overpic}[width=0.98\textwidth]{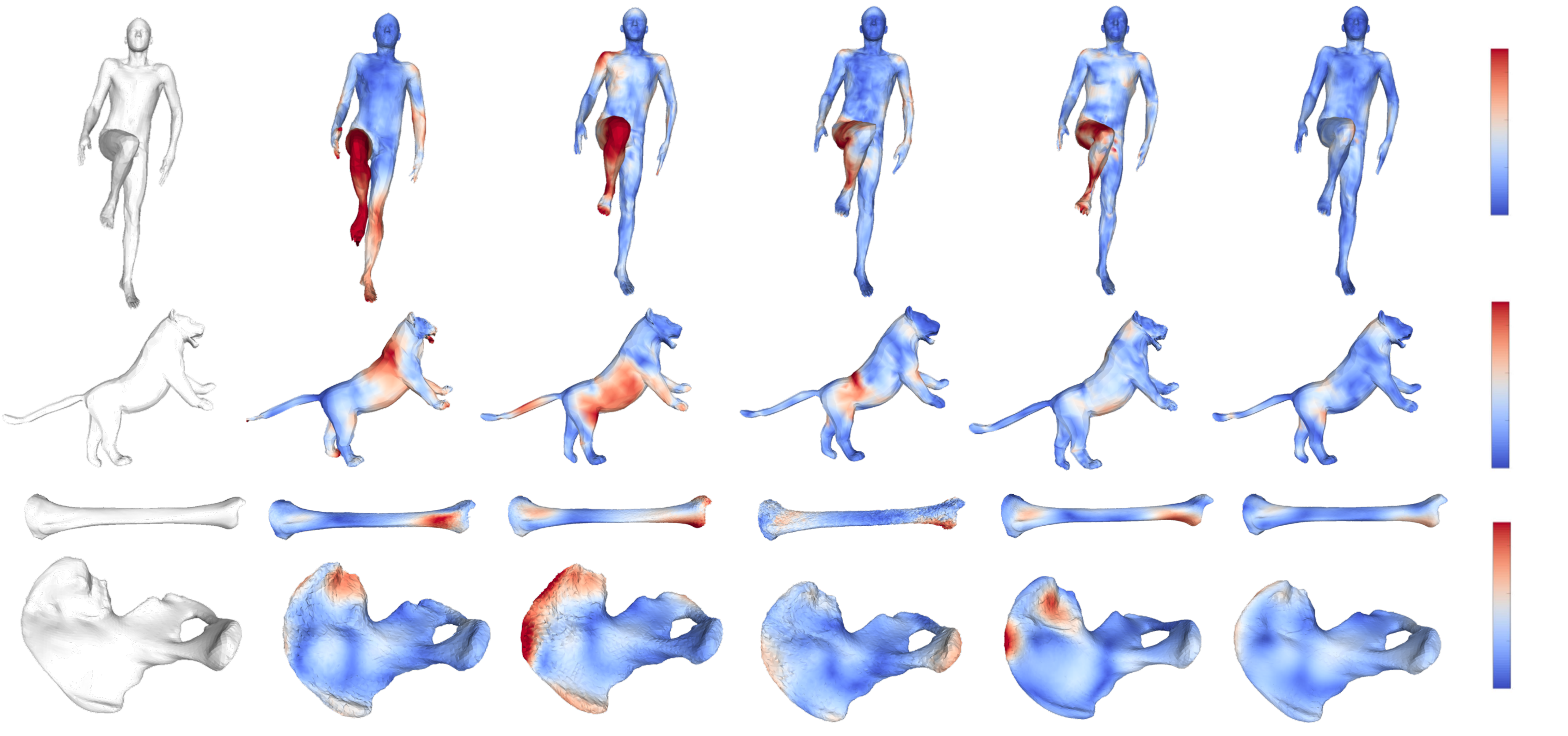}
\put(3,-0.8) {Ground truth}
\put(20,-0.8) {3DMM~\cite{bouritsas2019neural}}
\put(35,-0.8) {COMA~\cite{ranjan2018generating}}
\put(49,-0.8) {MeshConv~\cite{zhou2020fully}}
\put(66,-0.8) {Ours AD}
\put(78,-0.8) {Ours AD + ARAP}
\put(96.5, 43) {2cm}
\put(96.5,33.3) {0cm}
\put(96.5,25.5) {4cm}
\put(96.5,17.5) {0cm}
\put(96.5,12) {2cm}
\put(96.5,3) {0cm}
% \put(96.5,21.5) {0cm}
\end{overpic}
% \vspace{-0.3in}
\caption{Qualitative comparison of reconstruction results. We show results using the AD generator w/w.o ARAPReg.  Compared with baseline approaches, ours results with ARAPReg present less distortions and are locally smoother.}
\label{Figure:Corres:Based:Setting}
\end{figure*}
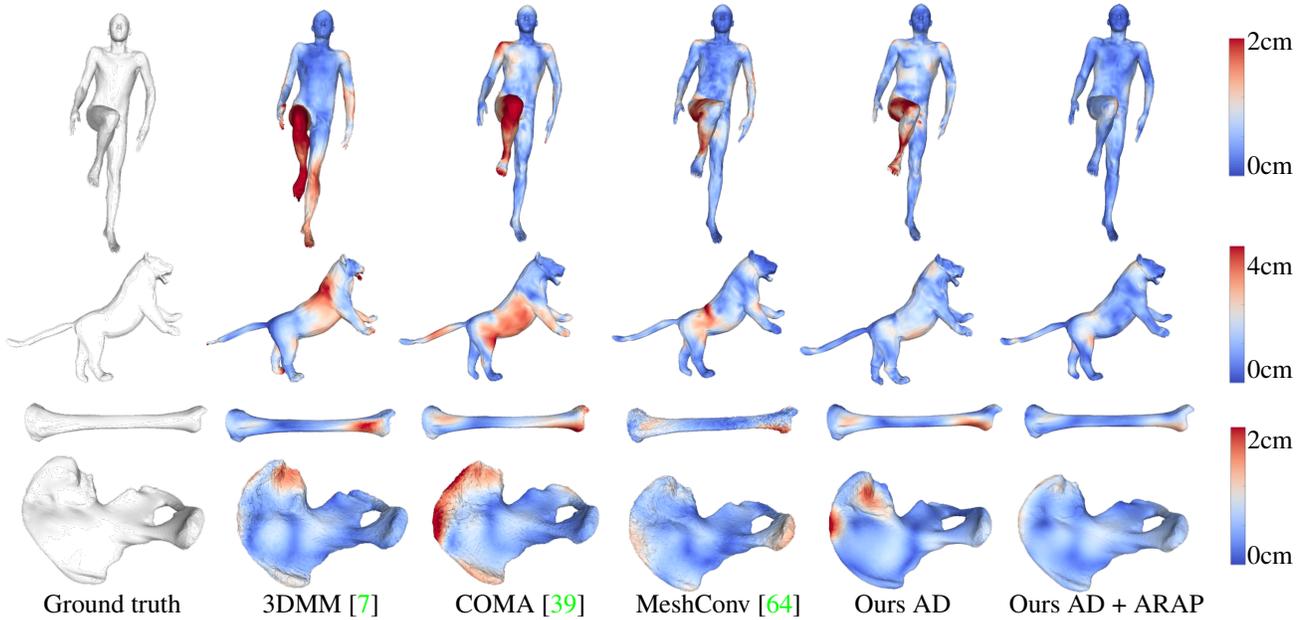

This section presents an experimental evaluation of ARAPReg. In Section~\ref{Subsec:Exp:Setup}, we present the experimental setup. We then analyze the experimental results in Section~\ref{Subsec:Results}. Finally, Section~\ref{Subsec:Abalation:Study} and Section~\ref{Subsec:Shape:Interpolation} describe an ablation study of the ASARReg loss and an evaluation of the shape interpolation application. Due to space issues, we defer more results and comparisons to the supp. material. %Please refer to the supp. material for more results. 

\subsection{Experimental Setup}
\label{Subsec:Exp:Setup}

% This section presents the experimental setup.

\noindent\textbf{Datasets.} The experimental evaluation considers three datasets: \textbf{DFAUST}~\cite{dfaust:CVPR:2017}, \textbf{SMAL}~\cite{zuffi20173d}, and \textbf{Bone}. The DFAUST dataset consists of 37,197 human shapes for training and 4,264 shapes for testing. All the shapes are generated using the SMPL model~\cite{Loper:2015:SMPL}. For the SMAL dataset, we randomly generate 400 shapes for following the shape sampling method in \cite{3dcoded}, where latent vectors are sampled from a normal distribution with zero mean and 0.2 standard deviations. We split them into 300 training shapes and 100 testing shapes.
The Bone dataset consists of four categories of real bones: Femur, Tibia, Pelvis, and Scapula, where each category has 40 training and 10 testing shapes. The consistent correspondences are obtained from interpolating landmark correspondences marked by experts.

\noindent\textbf{Baseline approaches.} We evaluate on four baselines: \textbf{SP-Disentangle \cite{ZhouBP20}}, \textbf{CoMA~\cite{ranjan2018generating}}, \textbf{ 3DMM~\cite{bouritsas2019neural}}, and \textbf{ MeshConv~\cite{zhou2020fully}}. 
They together represent the state-of-the-art results on learning mesh generators from a collection of meshes with dense correspondences. We evaluate the effectiveness of ARAPReg on these baselines and the absolute performance of our approach against these baselines.
%COMA~\cite{ranjan2018generating} and 3DMM~\cite{bouritsas2019neural} both combines a multi-resolution graph encoder and a graph decoder to reconstruct shapes. SP-Disentangle \cite{ZhouBP20} uses the model architecture of 3DMM~\cite{bouritsas2019neural} and add ARAP~\cite{Alexa:2000:ARAP} optimization to their output shapes as a post-processing step.  MeshConv~\cite{zhou2020fully} uses a fully-convolutional graph network to encode and decode shapes, which achieves the state-of-the-art quality.

\vspace{0.02in}

\noindent\textbf{Evaluation metrics.} Besides qualitative evaluations, we employ the reconstruction error metric (c.f.~\cite{ranjan2018generating,bouritsas2019neural,zhou2020fully}) for quantitative evaluations. Specifically, we compute the average per-vertex Euclidean distance for the input and reconstructed meshes. For VAE, the latent variable is given by the encoder . For AD, we optimize the latent variable to find the best reconstruction (c.f~\cite{YangLH18,zadeh2019variational}). The output shape is obtained by feeding the latent variable to the  decoder.

%\item \textsl{Symmetry error.} The symmetry error is a generic metric that applies to many organic and man-made shape collections. For Human, Animal, and Vase, we mark feature points on the base mesh and measure geodesic distortions between corresponding point pairs on synthetic shapes. This metric does not apply to asymmetric Bone shapes.   
%\end{enumerate}
%\begin{enumerate}
%\item \textsl{Self-intersection.} The percentage of triangles that are intersecting with other triangles. An ideal mesh generator should produce triangular meshes that are free of self-intersections. 
%\noindent\textbf{2. Shape Quality Measure.} The second metric assesses the local shape quality. Specifically, we study the distribution of the ratio between the shortest edge and the longest edge of each triangle. This metric assesses the meshing quality. Ratios of high-quality meshes should concentrate around one.
%\item \textsl{Vertex angle.} The distribution of vertex angles of all the triangles. This metric implicitly assesses the smoothness of the resulting mesh. The vertex angles of an ideal mesh concentrate around $\frac{\pi}{3}$.

\subsection{Analysis of Results}
\label{Subsec:Results}
\setlength\tabcolsep{4pt}
\begin{table}[t!]

 \begin{tabular}{ c |c  c c } 
 \hline
   & DFAUST & SMAL & Bone \\
 \hline
 SP-Disentangle. \cite{ZhouBP20} & 10.02 & 21.32 & 5.34 \\ 
 COMA\cite{ranjan2018generating} & 8.80 & 14.52 &  4.14\\ 
 3DMM\cite{bouritsas2019neural} & 7.39 & 17.78 &  4.03 \\
 MeshConv\cite{zhou2020fully} & 5.43 & 8.01 & 4.47\\
 \hline
 Ours-VAE (L1) & 5.45 & 9.11 & 4.09 \\
 Ours-VAE (L1 + ARAP) & 4.87 & 7.82 & 3.85 \\
 Ours-AD (L1) & 5.17 & 8.74 & 3.91\\
 Ours-AD (L1 + ARAP) & \textbf{4.52} & \textbf{6.68} & \textbf{3.76} \\
\end{tabular}
\caption{Correspondence-based MSE reconstruction error (mm) on test sets of DFaust, SMAL and Bone.}
\vspace{-0.1in}
\label{Table::corre-based}
\end{table}

\begin{figure*}
\centering
\begin{overpic}[width=1.0\textwidth]{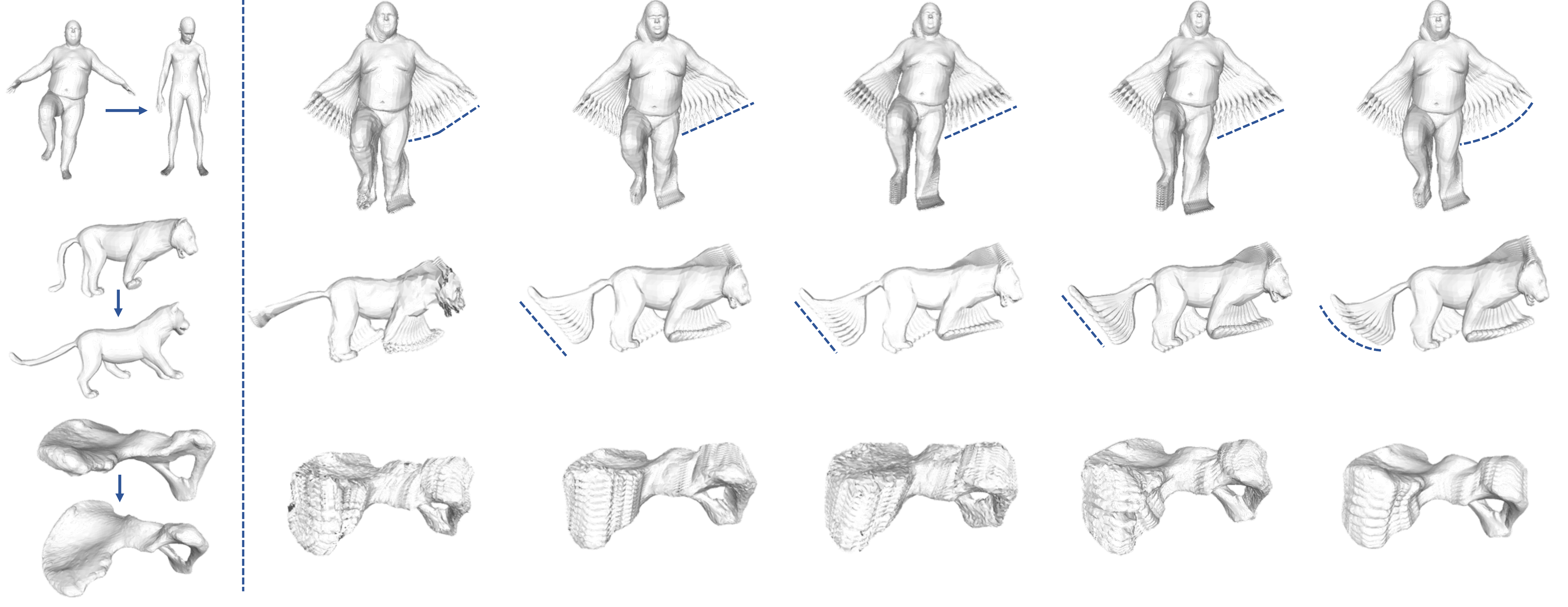}
\put(21,1) {3DMM~\cite{bouritsas2019neural}}
\put(37.5,1) {COMA~\cite{ranjan2018generating}}
\put(54,1) {MeshCov~\cite{zhou2020fully}}
\put(72,1) {Ours AD}
\put(84,1) {Ours AD + ARAP}
\end{overpic}
\vspace{-0.2in}
\caption{Interpolation results. The left column shows three groups of source and target shapes (connected by blue arrows). The remaining columns show ten intermediate shapes by linearly interpolating the latent codes of the source and target shapes. We show results using the AD generator w/w.o ARAPReg. Compared with baseline approaches, our results with ARAPReg show much smoother and more shape-preserving deformations.  }
\vspace{-0.2in}
\label{Figure:Shape:Interp:Comparison}
\end{figure*}

\begin{figure*}
\centering
\begin{overpic}[width=1.0\textwidth]{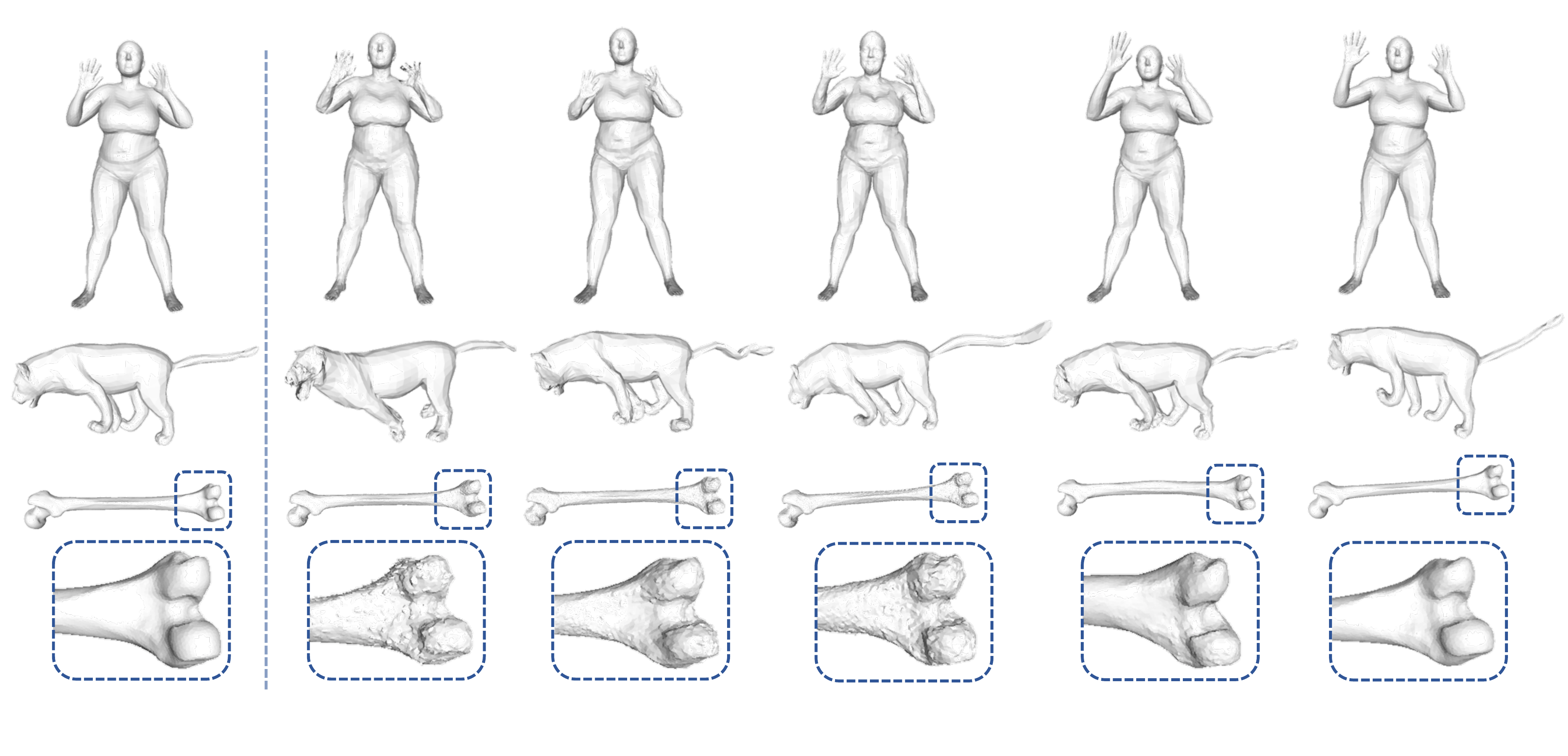}
\put(3.5,2.5) {Center Shape}
\put(21,2.5) {3DMM~\cite{bouritsas2019neural}}
\put(35.5,2.5) {COMA~\cite{ranjan2018generating}}
\put(50.5,2.5) {MeshConv~\cite{zhou2020fully}}
\put(70,2.5) {Ours AD}
\put(83,2.5) {Ours AD + ARAP}
\end{overpic}
\vspace{-0.3in}
\caption{Extrapolation results. Around one test center shape (left column), we randomly perturb its latent code $z$ within an Euclidean ball to generate perturbed shapes. We show results using the AD generator w/w.o ARAPReg. Our results with ARAPReg exhibit smooth and feature-preserving deformations. }
\label{Figure:Qualtative:Comparison}
\vspace{-0.1in}
\end{figure*}

\begin{figure}
\centering
\begin{overpic}[width=1.0\columnwidth]{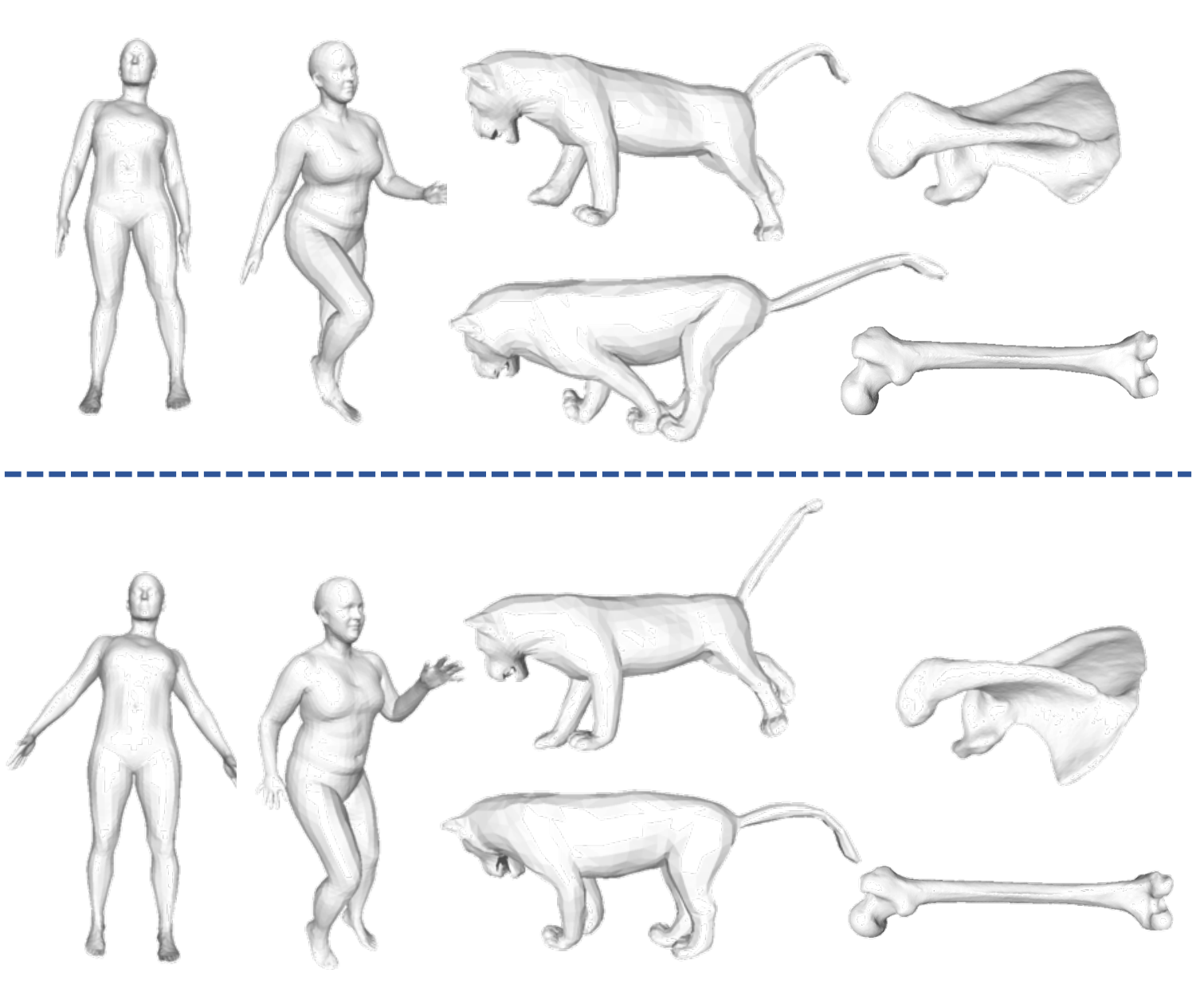}
\put(3.5,82.5) {Generated Shapes}
\put(3.5,37.5) {Closest Shapes}
\end{overpic}
\vspace{-0.2in}
\caption{Randomly generated shapes from our AD + ARAPReg framework and their closest shapes in the training set. Our network is able to generate reasonable shapes that are not in the training shape collection. Due to space constraints, results of our VAE framework are in the supp. material.}
\vspace{-0.2in}
\label{Figure:Shape:Closed}
\end{figure}

Table~\ref{Table::corre-based} compares our approach and baseline approaches in terms of the reconstruction error. Under the AD framework, our approach reduces the reconstruction error of baseline approaches by 16.8\%, 16.6\%, and 6.7\% on DFAUST, SMAL, and Bone, respectively. As the optimal latent-variable is optimized, AD framework achieves  better quality than VAE framework. 
%Note that SP-Disentangle\cite{ZhouBP20} also uses ARAP as a post-processing step and simply applies deformation between the output and the GT shape. Our method learns a shape space where local rigidity is preserved, which is helpful to generate novel shapes. 

Figure~\ref{Figure:Corres:Based:Setting} illustrates the reconstruction errors visually. Our approach improves from baseline approaches considerably. In particular, it improves from the top-performing approach MeshConv~\cite{zhou2020fully} at locations with large deformations (e.g., arms of humans) and non-rigid deformations (e.g., arms and torsos of humans). These improvements come from modeling the preservation of the local rigidity among neighboring shapes in the underlying shape space. Please refer to the supp. material for more results.

Figure~\ref{Figure:Shape:Closed} and the supp. material shows randomly generated shapes under our trained full VAE and AD models (i.e., with ARAPReg). We can see that the generated shapes nicely preserve important shape features such as fingers and faces of human shapes and tails of animal shapes. Moreover, the generated shapes are different from the closest training shape, indicating that the learned mesh generator has a strong generalization ability. 

{\small
\begin{table}[t!]
 \begin{tabular}{ c |c  c c } 
 \hline
   & 3DMM~\cite{bouritsas2019neural} & COMA~\cite{ranjan2018generating} &MeshConv~\cite{zhou2020fully} \\
 \hline
 No-Reg & 7.39 & 8.80 & 5.43 \\ 
 ARAPReg  & 6.72 & 4.87 & 5.02  \\
 \hline
\end{tabular}
\vspace{0.05in}
\caption{The effects of ARAPReg on different baselines on DFAUST dataset. The first row shows reported MSE reconstruction errors in their papers, and the second row shows results with ARAPReg on the same architecture with VAE training. ARAPReg achieves improvements on various baselines with different architectures. }
\vspace{-0.2in}
\label{Table::ARAP_baseline}
\end{table}
}
\subsection{Ablation Study}
\label{Subsec:Abalation:Study}

Table~\ref{Table::corre-based} shows our quantitative reconstruction results with and without ARAPReg under the VAE and AD settings. The effects of ARAPReg are salient. Under the AD setting, ARAPReg leads to 12.6\%,  23.5\%, and 4.5\% reductions of the reconstruction error on DFAUST, SMAL, Bone, respectively. 
Table~\ref{Table::ARAP_baseline} further shows the effect of ARAPReg  on various baselines under the VAE reconstruction pipeline. ARAPReg reduces the reconstruction error by building a better shape space that preserves the local rigidity constraint. 

%Moreover, ARAPReg helps the generator build a smooth shape space. Figure~\ref{Figure:Shape:Interp:Comparison} and  Figure~\ref{Figure:Qualtative:Comparison} illustrate the effects of ARAPReg in shape interpolation and extrapolation. After enforcing ARAPReg, the shape deformation becomes locally rigid and smooth. In other words, ARAPReg alleviates the issue of overfitting.
% Besides improving the reconstruction errors among regions that exhibit large pose changes (e.g., arms of humans), ARAPReg also leads to salient improvements around joints between different parts (e.g., between legs and torsos of humans and animals). One explanation is that these are the places where the local deformations are most salient among the training data. 
%Enforcing a local rigidity regularization can also alleviate the issue of overfitting. 

% Although the reductions in the Bone category are less significant in the reconstruction error, Figure~\ref{Figure:Qualtative:Comparison} and Figure~\ref{Figure:Shape:Interp:Comparison} show that the generated shapes when enforcing the ARAPReg loss are much cleaner than those without this loss.

\subsection{Shape Interpolation}
\label{Subsec:Shape:Interpolation}

We proceed to evaluate the effects of ARAPReg for the application of shape interpolation. Given two shapes $\bs{g}_1$ and $\bs{g}_2$, we first obtain their corresponding latent parameters $\bs{z}_1$ and $\bs{z}_2$. For the VAE model, $\bs{z}_i$ comes from the encoder. For the AD model, $\bs{z}_i$ comes from optimizing the reconstruction error. The interpolation is then done by linearly interpolating $\bs{z}_1$ and $\bs{z}_2$.

Figure~\ref{Figure:Shape:Interp:Comparison} compares our approach and baseline approaches on shape interpolation. We can see that our approach's interpolated shapes are smoother and more shape-preserving than those of the baseline approaches. Specifically, prominent shape features such as fingers are better preserved in our approach. Moreover, our approach introduces less distortion among joint regions. 

% Figure~\ref{Figure:Shape:Interp:Comparison}(Bottom) shows a result of interpolating two point clouds. In this case, we replace the vertex-based reconstruction loss with the sum of squared distances between nearest-neighbors (Please refer to the supp. material for details). Thanks to the learned mesh generator, the reconstructed shapes enhance shape features from the inputs. The interpolated shapes are again feature-preserving. 

\subsection{Shape Extrapolation}

We also evaluate the effects of ARAPReg for the application of shape extrapolation. Given a center shape $\bs{g}$, we first obtain its corresponding latent parameters $\bs{z}$. For the VAE model, $\bs{z}$ comes from the encoder. For the AD model, $\bs{z}$ comes from optimizing the reconstruction error. The extrapolation is then done by randomly sampling $\bs{\tilde{z}} \sim \bs{z} + \mathcal{N}(0, \sigma^2 S)$, where $S$ denotes scale for each latent dimension. We choose $\sigma = 0.2$ for all datasets.

Figure~\ref{Figure:Qualtative:Comparison} compares our approach and baseline approaches on shape extrapolation. We can see that our approach's generated shapes are smoother and more reasonable than baseline approaches in areas such as tails of animals, hands and arms of human.

%% file: 07_conclusions.tex
\section{Conclusions and Limitations}
\label{Section:Conclusions}

This paper introduces ARAPReg, an unsupervised loss functional for training shape generators. Experimental results show that enforcing this loss on meshed shape generators improves their performance. The resulting mesh generators produce novel generated shapes that are shape-preserving at multiple scales. 

ARAPReg has several limitations which can inspire future work. First, so far, ARAPReg only applies to training datasets with given correspondences. An interesting problem is to address unorganized shape collections that do not possess dense correspondences. Besides pre-computing correspondences, a promising direction is to explore the simultaneous learning  of the shape correspondences and the shape generator. Another limitation of ARAPReg is that it targets realistically deformable  shapes. Future directions are to study how to extend the formulation to handle synthetically generated shapes of any form and function.

\textbf{Acknowledgement.}  Chandrajit Bajaj would like to acknowledge the support from NIH-R01GM117594, by the Peter O’Donnell Foundation, and in part from a grant from the Army Research Office accomplished under Cooperative Agreement Number W911NF-19-2-0333. Junfeng Jiang would like to acknowledge support from Jiangsu NSF under Grant BK20181158 and NSF of China (NSFC) under Grant 61772172. Qixing Huang is supported by NSF Career IIS-2047677, NSF HDR TRIPODS-1934932, and Gifts from Wormpex AI Research and Snap Inc.

%% file: 11_supp.tex
\section{More Quantitative Results}
\subsection{Ablation Study on Pose and Shape Variation in Section 4.3}
In the Section 4.2, we introduced decoupling shape and pose variations to improve ARAPReg. Here we show an ablation study of this decoupling. In Table.\ref{supp::Table::shape_pose}, we show MSE reconstruction error in AD framework w/w.o shape and pose decoupling. Specifically, in the non-decoupling setting, we use the L2 formulation in Proposition 2, where all directions are penalized equally. 

\subsection{Comparison with ARAP deformation from the base mesh}
Here we show the comparison between our method and the traditional ARAP deformation method, where an ARAP deformation is applied between the base mesh and the output mesh for regularization (c.f.~\cite{DBLP:conf/cvpr/HabermannXZPT20,DBLP:conf/nips/LiLMKW0K20,ZhouBP20}). 
In Table~\ref{supp::Table::arap_deform}, we show results on DFAUST and SMAL datasets. On DFAUST dataset, there are large deformations among the underlying shapes, and the approach of enforcing an ARAP loss to the base shape is significantly worse than without the ARAP loss. In the SMAL dataset, we pick all samples with the same shape but different poses, the ARAP loss to the base shape offers slight performance gains. However, ARAPReg still outperforms this simple baseline considerably.

\begin{table}[t!]

 \begin{tabular}{ c |c  c c } 
 \hline
  & DFAUST & SMAL & Bone \\
   \hline

 W.o. Decoupling & 4.90 & 7.23 & 3.82 \\
 With Decoupling & \textbf{4.52} & \textbf{6.68} & \textbf{3.76} \\
  \hline

\end{tabular}
\caption{Ablation study on shape and pose variation. In w.o. decoupling setting, all directions are penalized equally. With decoupling setting is the setting in the main paper, where pose directions are penalized more than shape directions.}
\vspace{-0.1in}
\label{supp::Table::shape_pose}
\end{table}
\begin{table}[t!]
 \begin{tabular}{ c |c  c c } 
 \hline
  & DFAUST & SMAL  \\
   \hline
   No ARAP  & 5.17 & 8.74 \\
 ARAP Deform. & 24.55 & 7.67  \\
 Ours & \textbf{4.52} & \textbf{6.68} &  \\
  \hline
\end{tabular}
\caption{Comparison between our method and the traditional ARAP deformation method. We show reconstruction errors of AD model without ARAP, with traditional ARAP and with our method. The traditional method couldn't handle large pose variation and shape distortion. }
\vspace{-0.1in}
\label{supp::Table::arap_deform}
\end{table}
\section{More Implementation Details}
\subsection{Model Architecture}

Our VAE model consists of a shape encoder and a decoder. Our AD model only contains a decoder. Both encoder and decoder are composed of Chebyshev convolutional filters with $K = 6$ Chebyshev polynomials \cite{ranjan2018generating}.The VAE model architecture is based on \cite{ranjan2018generating}. We sample 4 resolutions of the mesh connections of the template mesh. The encoder is stacked by 4 blocks of  convolution + down-sampling layers. The decoder is stacked by 4 blocks of convolution + up-sampling layers. There's two fully connected layers connecting the encoder, latent variable and the decoder. For the full details, please refer to our Github repository. 

\subsection{Reconstruction evaluation}
In the AD model, there's no shape encoder to produce latent variables so we add an in-loop training process to optimize shape latent variables, where we freeze the decoder parameters and  optimize latent variables for each test shape. 
In the VAE training, we also add some refinement steps on the latent variable optimization where we freeze the decoder. We apply this refinement step to both methods w/w.o ARAPReg. 

\section{More Results}

\begin{figure*}
\centering
\begin{overpic}[width=0.8\textwidth]{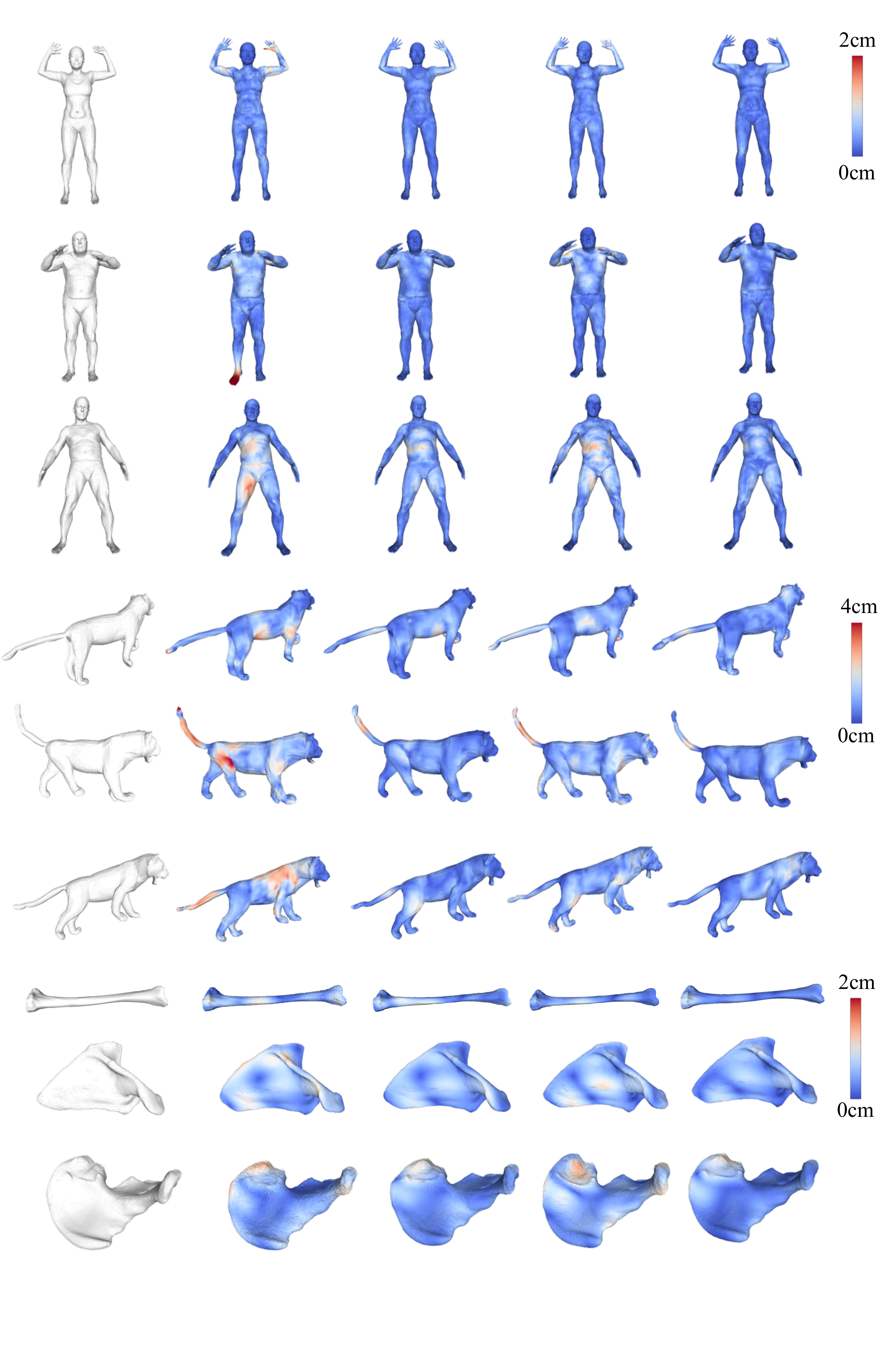}
\put(4,4) {Ground truth}

\put(19,4) {VAE}
\put(29,4) {VAE+ARAP}
\put(43,4) {AD}
\put(51,4) {AD+ARAP}

% \put(36,3.5) {COMA~\cite{ranjan2018generating}}
% \put(50,3.5) {MeshConv~\cite{zhou2020fully}}
% \put(67,3.5) {Ours AD}
% \put(80,3.5) {Ours AD + ARAP}
% \put(95.5,52) {2cm}
% \put(95.5,38.3) {0cm}
% \put(95.5,35.5) {4cm}
% \put(95.5,22.5) {0cm}
% \put(95.5,20) {2cm}
% \put(95.5,7) {0cm}
% \put(96.5,21.5) {0cm}
\end{overpic}
\vspace{-0.3in}
\caption{More qualitative results of reconstruction. We show results using  VAE and AD generator w/w.o ARAPReg.}
\label{Figure:supp::more_recons}
\end{figure*}

\begin{figure*}
\centering
\begin{overpic}[width=0.8\textwidth]{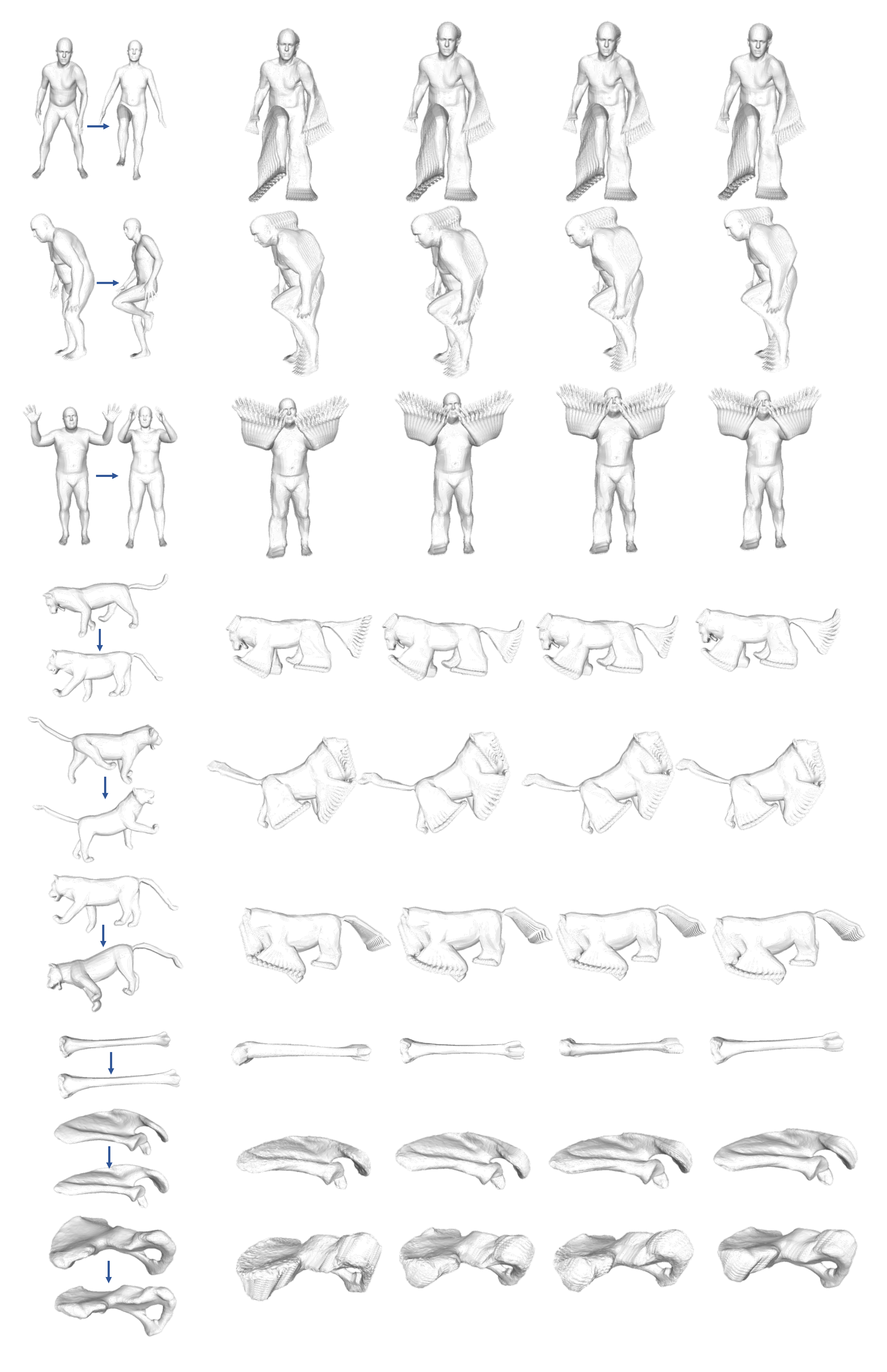}
\put(20.5,2) {VAE}
\put(30,2) {VAE+ARAP}
\put(44,2) {AD}
\put(53.5,2) {AD+ARAP}

\end{overpic}
\vspace{0.1in}
\caption{More interpolation results. We show results using  VAE and AD generator w/w.o ARAPReg.}
\label{Figure:supp::more_interp}
\end{figure*}

\begin{figure*}
\centering
\begin{overpic}[width=0.8\textwidth]{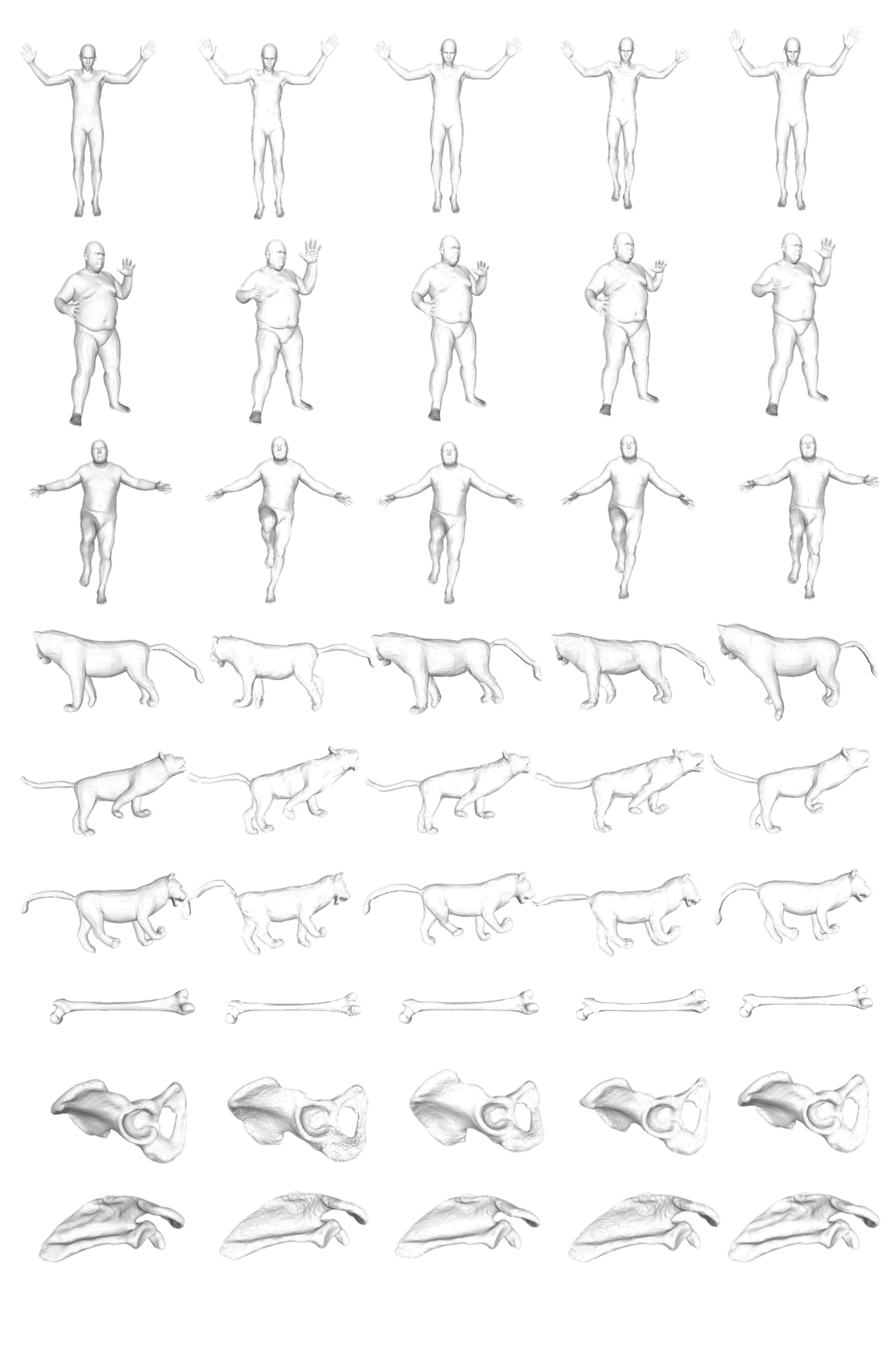}
\put(4,4) {Center Shape}
\put(20,4) {VAE}
\put(30,4) {VAE+ARAP}
\put(46,4) {AD}
\put(55,4) {AD+ARAP}
\end{overpic}
\vspace{-0.3in}
\caption{More extrapolation results. We show results using  VAE and AD generator w/w.o ARAPReg.}
\label{Figure:supp::more_extrap}
\end{figure*}

\begin{figure*}
\centering
\begin{overpic}[width=0.8\textwidth]{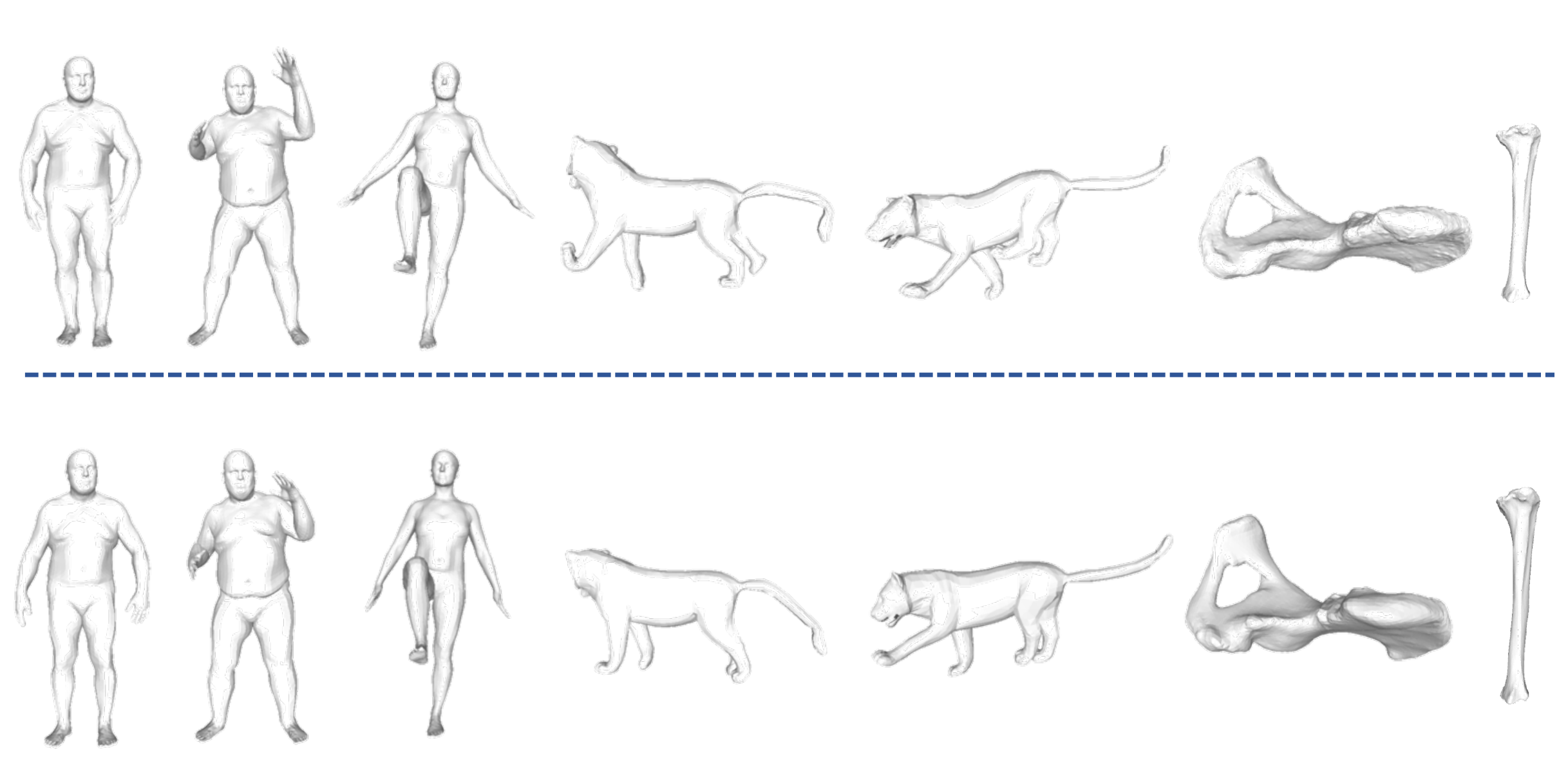}
\put(4,48) {Generated Shapes}
\put(4,22) {Closest Shapes}
\end{overpic}

\caption{Randomly generated shapes from our VAE frame work and their closed shapes in the training set.}
\label{Figure:supp::more_close}
\end{figure*}

In this section, we show more results of reconstruction (Fig.\ref{Figure:supp::more_recons}), interpolation (Fig.\ref{Figure:supp::more_interp}) and extrapolation (Fig.\ref{Figure:supp::more_extrap}) of our methods in variational auto-encoder (VAE) and auto-decoder (AD) frameworks, with and without ARAPReg. We also show more closest shapes for randomly generated shapes in VAE framework with ARAPReg in Fig. \ref{Figure:supp::more_close}.

%% file: 08_app_regu_term.tex
\section{Proofs of Propositions in Section 4.2}

\subsection{Proof of Prop.1}
\label{App:Proof:Prop:Rigidity}
For a shape $\textbf{g} \in \set{R}^{3n}$ with an infinitesimal vertex displacement $\textbf{x} \in \set{R}^{3n}$ and $\|\textbf{x}\|_2 \leq \epsilon$, the local rigidity energy is
\begin{equation}
    E(\textbf{g}, \textbf{x}) = \min_{\{A_i \in SO(3)\}} \sum_{(i,j) \in \set{E}} w_{ij} \|(A_i-I_3) (\textbf{g}_i - \textbf{g}_j) - (\textbf{x}_i-\textbf{x}_j) \|^2
\end{equation}
where $A_i$ is a 3D rotation matrix denoting the local rotation from $\textbf{g}_i - \textbf{g}_j$ to $(\textbf{g}_i + \textbf{x}_i) - (\textbf{g}_j + \textbf{x}_j)$. Note that here vector indexing is vertex indexing, where $\textbf{g}_i = \textbf{g}_{3i:3(i+1)}$.

Since the zero and first-order derivatives from $E$ to $\textbf{x}$ around zero is 0:
\begin{equation}
    E(\textbf{g}, \textbf{x})|_{\textbf{x}=\textbf{0}} = 0, \quad \frac{\partial E(\textbf{g}, \textbf{x})}{\partial \textbf{x}}|_{\textbf{x}=\textbf{0}} = \textbf{0}
\end{equation}

We can use second-order Taylor expansion to approximate the energy $E$ when $\textbf{x}$ is around zero:
\begin{equation}
    E(\textbf{g}, \textbf{x}) \approx \frac{1}{2} \textbf{x}^T \frac{\partial^2 E}{\partial \textbf{x}^2} \textbf{x}
\end{equation}

\begin{proposition}
Given a function $g(\textbf{x}) = \min_\textbf{y} f(\textbf{x}, \textbf{y})$, and define $\textbf{y}(\textbf{x}) = \text(argmin)_\textbf{y} f(\textbf{x},\textbf{y})$ such that $g(\textbf{x}) = f(\textbf{x}, \textbf{y}(\textbf{x}))$, 
\begin{equation}
    \frac{\partial^2 g}{\partial \textbf{x}^2} = \frac{\partial^2 f}{\partial \textbf{x}^2} -   \frac{\partial^2 f}{\partial \textbf{x} \partial \textbf{y}} 
    (\frac{\partial^2 f}{\partial \textbf{y}^2})^{-1}
    \frac{\partial^2 f}{\partial \textbf{y} \partial \textbf{x}}
\end{equation}
\label{Prop:app_hessian}
\end{proposition}

By treating each $A_i$ as a function of $\textbf{x}$, we can rewrite our energy as 
\begin{equation}
    E(\textbf{g}, \textbf{x}) = f_{\textbf{g}}(\textbf{x}, A(\textbf{x}))
\end{equation}
where $A$ is the collection of all $A_i$. 

By using Prop.\ref{Prop:app_hessian}, we can get the Hessian from $E$ to $\textbf{x}$. 

In the above formulation,  $A_i$ is in the implicit form of \textbf{x}. Now we use Rodrigues' rotation formula to write is explicitly. 
For a rotation around an unit axis $k$ with an angle $\theta$, its rotation matrix is 
\begin{equation}
    A_i = I + \sin_\theta \textbf{k}\times + (1-\cos_\theta) (\textbf{k}\times)^2
    \label{Eq:app:rotation}
\end{equation}
where $\textbf{k}\times$ is the cross product matrix of vector $\textbf{k}$. 

Since here we apply infinitesimal vertex displacement, rotation angle $\theta$ is also infinitesimal. We can approximate \ref{Eq:app:rotation} as
\begin{equation}
    A_i \approx I + \theta \textbf{k}\times + \frac{1}{2} (\theta \textbf{k}\times)^2
    \label{Eq:app:rotation2}
\end{equation}
Let $\textbf{c} = \theta \textbf{k}$ and only preserve the first two terms:

\begin{align}
    E(\textbf{x}) &\approx \min_{\{\textbf{c}_i\}} \sum_{(i,j) \in \set{E}} w_{ij} \|\textbf{c}_i \times \textbf{e}_{ij} - (\textbf{x}_i-\textbf{x}_j) \|^2\\
    &= \min_{\{\textbf{c}_i\}} \sum_{(i,j) \in \set{E}} w_{ij} \|\textbf{e}_{ij} \times \textbf{c}_i + (\textbf{x}_i-\textbf{x}_j) \|^2
\end{align}
where $\textbf{e}_{ij} = \textbf{p}_i - \textbf{p}_j$

From Prop. \ref{Prop:app_hessian}, we can compute the hessian from $E$ to $\textbf{x}$ by writing $E(\textbf{g}, \textbf{x}) = f_{\textbf{g}}(\textbf{x}, \textbf{c}(\textbf{x}))$.

We rewrite our energy function in matrix form
\begin{equation}
    E =
    \begin{bmatrix}
    \textbf{x}^T & \textbf{c}^T 
    \end{bmatrix} 
    \begin{pmatrix}
    L \otimes I_3 & B \\
    B^T & C \\
    \end{pmatrix}
    \begin{bmatrix}
    \textbf{x} \\
    \textbf{c} 
    \end{bmatrix} 
\end{equation}
where $\otimes$ denotes the kronecker product or tensor product. 

The Hessian from $E$ to $\textbf{x}$ around zero is
\begin{equation}
    H_R(\textbf{g}) = L \otimes I_3 - B^T C^{-1} B
\end{equation}

Now we compute each term of $H_R(\textbf{g})$.
Expand $f_{\textbf{g}}(\textbf{x}, \textbf{c}(\textbf{x}))$:
\begin{align*}
    f(\textbf{x}, \textbf{c}(\textbf{x})) &= \sum_{(i,j) \in \set{E}} w_{ij} \|\textbf{e}_{ij} \times \textbf{c}_i + (\textbf{x}_i-\textbf{x}_j) \|^2 \\
    &= \sum_{(i,j) \in \set{E}} w_{ij} (\textbf{x}_i^2 + \textbf{x}_j^2 - 2 \textbf{x}_i \textbf{x}_j +
    2(\textbf{e}_{ij} \times \textbf{c}_i)^T (\textbf{x}_i-\textbf{x}_j) \\
    & + (\textbf{e}_{ij} \times \textbf{c}_i)^T (\textbf{e}_{ij} \times \textbf{c}_i))
\end{align*}

% The $3 \times 3$ block second-order derivatives are: 
% \begin{align}
%     (\frac{\partial^2 f}{\partial \textbf{x}^2})_{ii} &= 2  \sum_{j \in \set{N}_i} w_{ij} \\
%     (\frac{\partial^2 f}{\partial \textbf{x}^2})_{ij} &= -2 w_{ij} \quad (i \neq j) \\
%     (\frac{\partial^2 f}{\partial \textbf{c} \partial \textbf{x}})_{ii} &= 2  \sum_{j \in \set{N}_i} w_{ij} \textbf{e}_{ij} \times \\
%      (\frac{\partial^2 f}{\partial \textbf{c} \partial \textbf{x}})_{ij} &= -2 w_{ij}\textbf{e}_{ij} \times  \quad (i \neq j)\\
%      (\frac{\partial^2 f}{\partial \textbf{c}^2})_{ii} &= 2  \sum_{j \in \set{N}_i} w_{ij} (\textbf{e}_{ij} \times)^T (\textbf{e}_{ij} \times)
% \end{align}
% where $\textbf{e}_{ij} \times$ denotes the cross product matrix for vector $\textbf{e}_{ij}$. 

$L$ is the weighted graph Laplacian,
\begin{equation}
    \textbf{L}_{ij}=\left\{
                \begin{array}{ll}
                   \sum_{k \in \set{N}_i} w_{ik}, \quad i=j\\
                   -w_{ij}, \quad i\neq j \text{and} (i,j)\in \set{E} \\
                   0, \quad otherwise
                \end{array}
              \right.
\end{equation}

The matrix $\textbf{B}$ is a block matrix whose $3 \times 3$ blocks are defined as
\begin{equation}
    B_{ij}=\left\{
                \begin{array}{ll}
                   \sum_{k \in \set{N}_i} w_{ik} \textbf{e}_{ik} \times, \quad i=j\\
                   -w_{ij} \textbf{e}_{ij} \times, \quad i\neq j, (i,j)\in \set{E} \\
                   0, \quad otherwise
                \end{array}
              \right.
\end{equation}

Finally,C = diag$(C_1 ...C_{|P|})$ is a block diagonal matrix
\begin{align}
    C_i &= \sum_{j \in \set{N}_i} w_{ij} (\textbf{e}_{ij} \times)^T (\textbf{e}_{ij} \times)\\
    &=\sum_{j \in \set{N}_i} w_{ij} \|\textbf{e}_{ij}\|^2_2 \textbf{I}_3 - \textbf{e}_{ij} \textbf{e}_{ij}^T
\end{align}
% Because
% \begin{align}
%     (\textbf{e} \times)^T (\textbf{e} \times) &= \begin{bmatrix}
%     0 & e_3 & -e_2 \\
%     -e_3 & 0 & e_1 \\
%     e_2 & -e_1 & 0 
%     \end{bmatrix}
%     \begin{bmatrix}
%     0 & -e_3 & e_2 \\
%     e_3 & 0 & -e_1 \\
%     -e_2 & e_1 & 0 
%     \end{bmatrix} \\
%     &= \begin{bmatrix}
%     e_2^2 + e_3^2 & -e_1 e_2 & -e_2 e_3 \\
%     -e_2 e_1 & e_1^2 + e_3^2 & -e_2 e_3 \\
%     -e_3 e_1 & -e_3 e_2 &  e_1^2 + e_2^2
%     \end{bmatrix}\\
%     &= \|e\|^2_2\textbf{I}_3 - \textbf{e} \textbf{e}^T
% \end{align}

% So 
% \begin{equation}
%     \textbf{C}_i = \sum_{j \in \set{N}_i} w_{ij} \|\textbf{e}_{ij}\|^2_2 \textbf{I}_3 - \textbf{e}_{ij} \textbf{e}_{ij}^T
% \end{equation}

% Now given a vector $\textbf{a}$, the quadratic form of the above $H_R(g)$ is
% \begin{align*}
%     \textbf{a}^T H_R(g)\textbf{a} &= \sum_i [\sum_{j\in \set{N}_i} w_{ij} \|\textbf{a}_i - \textbf{a}_j\|^2_2 - (\sum_{j\in \set{N}_i} \textbf{e}_{ij} \times (\textbf{a}_i - \textbf{a}_j))^T \\
%     &(\sum_{j\in \set{N}_i} \|\textbf{e}_{ij}\|^2_2 \textbf{I}_3 - \textbf{e}_{ij} \textbf{e}_{ij}^T)^{-1}(\sum_{j\in \set{N}_i} \textbf{e}_{ij} \times (\textbf{a}_i - \textbf{a}_j)) ]
% \end{align*}

which ends the proof.
\qed

\subsection{Proof of Prop.2}
\label{Proof:Prop:Local:Rigidity:Expression}

Consider the eigen-decomposition of 
$$
\overline{H}_{R}(\bs{g},J): = U \Lambda U^T,
$$
where 
$$
\Lambda = \diag\big(\lambda_1(\overline{H}_{R}(\bs{g},J)),\cdots, \lambda_k(\overline{H}_{R}(\bs{g},J))\big).
$$
Let $\overline{y} = U^T\bs{y}$. Then
\begin{align*}
\int_{\bs{y}}\bs{y}^T \overline{H}_{R}(\bs{g},J)\bs{y} & = \int_{\bs{y}}\overline{\bs{y}}^T \Lambda \overline{\bs{y}} =\int_{\bs{y}} \sum\limits_{i=1}^{k}\lambda_i(\overline{H}_{R}(\bs{g},J))\overline{y}_i^2 \\
& = \sum\limits_{i=1}^{k}\lambda_i(\overline{H}_{R}(\bs{g},J))\int_{\overline{\bs{y}}}\overline{\bs{y}}_i^2 d\overline{\bs{y}} \\
& = \frac{1}{k}\sum\limits_{i=1}^{k}\lambda_i(\overline{H}_{R}(\bs{g},J))\int_{\overline{\bs{y}}}\sum\limits_{i=1}^{k}\overline{\bs{y}}_i^2 d\overline{\bs{y}} \\
& = \frac{\textup{Vol}(S^k)}{k}\sum\limits_{i=1}^{k}\lambda_i(\overline{H}_{R}(\bs{g},J).
\end{align*}
\qed

% \qed

% \subsection{Proof of Prop.2 in Section 4.2}
% \label{Proof:Prop:Local:Rigidity:Gradient:Expression}

% The proof is a direct application of
% $$
% d\lambda_i(A) = \bs{u}_i(A)^TdA \bs{u}_i(A)
% $$
% where $\lambda_i(A)$ and $\bs{u}_i(A)$ are the $i$-th eigenvalue and eigenvector of a symmetric matrix A. 
% \qed 

%% file: 10_app_gradients.tex
\section{Gradient of Loss Terms}
\label{Section:Derivative:Computation}

%This section presents gradients of various loss terms with respect to the output of the generator $\bs{g}^{\theta}(\bs{z}_i)$ and the latent variables $\bs{z}_i$. 

This section presents the gradients of the loss to the rigidity term.

For simplicity, we will express formulas for gradient computation using differentials. Moreover, we will again replace $\bs{g}^{\theta}$ and $\frac{\partial \bs{g}^{\theta}}{\partial \bs{z}}(\bs{z})$ with $\bs{g}$ and $J$ whenever it is possible. The following proposition relates the differential of $r_R(\bs{g},J)$ with that of $\overline{H}_{R}(\bs{g},J))$.  
\begin{proposition}
\begin{equation}
d r_R(\bs{g},J) = \alpha\sum\limits_{i=1}^{k}\frac{\bs{u}_i^Td(\overline{H}_{R}(\bs{g},J))\bs{u}_i}{\lambda_i^{1-\alpha}(\overline{H}_{R}(\bs{g},J))}.
\label{Eq:Hessian:Derivative}
\end{equation}
Recall that $\lambda_i$ and $\bs{u}_i$ are eigenvalues of eigenvectors of $\overline{H}_{R}(\bs{g},J))$.
\label{Prop:Local:Rigidity:Gradient:Expression}
\end{proposition}
\noindent\textsl{Proof:} 
The proof is straight-forward using the gradient of the eigenvalues of a matrix, i.e., 
$$
d\lambda = \bs{u}^TdH\bs{u}
$$
where $\bs{u}$ is the eigenvector of $H$ with eigenvalue $\lambda$. The rest of the proof follows from the chain rule. 
\qed

We proceed to describe the explicit formula for computing the derivatives of $\bs{u}_i^Td(\overline{H}_{R}(\bs{g},J))\bs{u}_i$. First of all, applying the chain rule leads to
\begin{align*}
& \ \ \bs{u}_i^Td(\overline{H}_{R}(\bs{g},J))\bs{u}_i = 2\Big((J\bs{u}_i)^TH_{R}(\bs{g})(dJ\cdot \bs{u}_i)\nonumber \\
  & - (A(\bs{g}) J\bs{u}_i)^T D(\bs{g})^{-1}\cdot \big(dA(\bs{g})\cdot (J\bs{u}_i)\big)\Big) \nonumber \\
 &\ \ + \big(D(\bs{g})^{-1}A(\bs{g})J\bs{u}_i\big)^T dD(\bs{g}) \big(D(\bs{g})^{-1}A(\bs{g})J\bs{u}_i\big).
\label{Eq:Derivative}    
\end{align*}

It remains to develop formulas for computing $dJ\cdot \bs{u}_i$, $dA(\bs{g})\cdot (J\bs{u}_i)$, and $dD(\bs{g})$. Note that $J = \frac{\partial \bs{g}^{\theta}}{\partial \bs{z}}(\bs{z})$. We use numerical gradients to compute $dJ\cdot \bs{u}_i$, which avoid computing costly second derivatives of the generator:
\begin{equation}
d(\frac{\partial \bs{g}^{\theta}}{\partial \bs{z}}(\bs{z}))\cdot \bs{u}_i \approx \sum\limits_{l=1}^{k} u_{il} (d\bs{g}^{\theta}(\bs{z} + s\bs{e}_l)-d\bs{g}^{\theta}(\bs{z}))
\label{Eq:Numerical:Gradient}    
\end{equation}
where $s = 0.05$ is the same hyper-parameter used in defining the generator smoothness term; $\bs{e}_l$ is the $l$-th canonical basis of $\R^k$; $u_{il}$ is the $l$-th element of $\bs{u}_i$. 

The following proposition provides the formulas for computing the derivatives that involve $A(\bs{g})$ and $D(\bs{g})$.
\begin{proposition}
\begin{align}
& dA(\bs{g})\cdot (J\bs{u}_i) = -  \set{A}(J\bs{u}_i)\cdot d\bs{g} \nonumber \\
& \bs{c}^TdD(\bs{g})\cdot \bs{c}  = 2\sum\limits_{i=1}^{n}\sum\limits_{k\in \set{N}(i)}\Big((\bs{g}_i-\bs{g}_k)^T(d\bs{g}_i-d\bs{g}_k)\|\bs{c}_i\|^2 \nonumber \\
& \quad -\big(\bs{c}_i^T(d\bs{g}_i-d\bs{g}_k)\big)\cdot \big((\bs{g}_i-\bs{g}_k)^T\bs{c}_i\big)\Big)    
\end{align}
\end{proposition}
\noindent\textsl{Proof:} 

(1). $dA(\bs{g})\cdot (J\bs{u}_i)$:

Let's denote $J\bs{u}_i$ as $\bs{a}$. Now we prove $ (A(\bs{g})\cdot \bs{a}) = (A(\bs{a})\cdot \bs{g})$. Then we will have $d(A(\bs{g}))v J\bs{u}_i = d(A(\bs{g})\cdot J\bs{u}_i) =d(A(J\bs{u}_i)\cdot \bs{g}) = A(J\bs{u}_i)\cdot d(\bs{g})$.

\begin{align*}
    & (A(\bs{g})\bs{a})_i = \sum_j A_{ij}(\bs{g}) \bs{a}_j \\
    & = \sum_{k \in N(i)} \bs{v}_ik \times (\bs{a}_i - \bs{a}_k) = \sum_{k \in N(i)} \bs{v}_ik \times \bs{a}_ik \\
    & = - \sum_{k \in N(i)} \bs{a}_ik \times \bs{v}_ik = \sum_j A_{ij}(\bs{a}) \bs{g}_j \\
    & = (A(\bs{a})\bs{g})_i
\end{align*}
This finishes the proof. 

(2). $\bs{c}^TdD(\bs{g})\cdot \bs{c}$:

We have $\bs{c}_i^T D_{ii}(\bs{g}) \cdot \bs{c}_i = \sum_{k \in N(i)} (\|\bs{v}_{ik}\|^2 \|\bs{c}_i\|^2 -  \bs{c}_i^T \bs{v}_{ik} \bs{v}_{ik}^T \cdot \bs{c}_i)$. We only need to compute the gradient of $\|\bs{v}_{ik}\|^2$ and $\bs{v}_{ik} \bs{v}_{ik}^T $. Note that $\|\bs{v}_{ik}\|^2 = \bs{v}_{ik}^T \bs{v}_{ik}$. 

For a vector $\bs{a}$, we have $d(\bs{a}^T \bs{a})= d(\bs{a}^T) \bs{a} + \bs{a}^T d(\bs{a}) =  d(\bs{a})^T \bs{a} + \bs{a}^T d(\bs{a}) = 2 \bs{a}^T d(\bs{a})$ and similarly, $d(\bs{a} \bs{a}^T) = 2 d(\bs{a}) \bs{a}^T$. We use these two results to our derivation and we will get the results above. 

\begin{align*}
    & \bs{c}^TdD(\bs{g})\cdot \bs{c}  \\
   & = \sum_i \sum_{k \in N(i)} (d(\|\bs{v}_{ik}\|^2) \|\bs{c}_i\|^2 -  \bs{c}_i^T d(\bs{v}_{ik} \bs{v}_{ik}^T )\cdot \bs{c}_i)\\
    & = \sum_i \sum_{k \in N(i)} (d(\bs{v}_{ik}^T \bs{v}_{ik}) \|\bs{c}_i\|^2 -  \bs{c}_i^T d(\bs{v}_{ik} \bs{v}_{ik}^T )\cdot \bs{c}_i)\\
    &= \sum_i \sum_{k \in N(i)} 2\Big((\bs{g}_i-\bs{g}_k)^T(d\bs{g}_i-d\bs{g}_k)\|\bs{c}_i\|^2 \nonumber \\
&\quad -\big(\bs{c}_i^T(d\bs{g}_i-d\bs{g}_k)\big)\cdot \big((\bs{g}_i-\bs{g}_k)^T\bs{c}_i\big)\Big)
\end{align*}

%\noindent\textsl{Proof:} Please refer to the supp. material. \qed

% \subsection{Derivatives of Spectral Embedding}
% \label{Subsec:Descriptor:Based:Gradients}

% The derivatives of the spectral embedding are given by 
% $$
% d(\frac{\bs{f}_i}{\sqrt{\lambda_i}}) = \frac{d\bs{f}_i}{\sqrt{\lambda_i}} - \frac{1}{2}\frac{\bs{f}_id\lambda_i}{\sqrt{\lambda_i^3}}.
% $$
% The following two formulas provide the derivatives of $\lambda_i$ and $\bs{f}_i$:
% \begin{align}
% d\lambda_i & = \bs{f}_i dL \bs{f}_i, \nonumber \\
% d\bs{f}_i  & = \sum\limits_{j\neq i}\frac{1}{\lambda_j-\lambda_i} (\bs{f}_i^T\bs{f}_j)\bs{f}_j. 
% \label{Eq:B:3}
% \end{align}
% Our implementations use the reduced eigen-decomposition to compute (\ref{Eq:B:3}). 

% It remains to compute $dL$ with respect to $\bs{g}^{\theta}$, which is reduces to computing the derivatives of cotangent of a triangle angle with respect to the three vertices. This can be achieved by applying the chain rule. 

%\subsection{Smoothness Term}
\label{Subsec:Smoothness:Term:Gradients}